\documentclass[letterpaper]{article}

\usepackage{aaai2026}
\nocopyright
\usepackage{times}
\usepackage{helvet}
\usepackage{courier}
\usepackage[hyphens]{url}
\usepackage{graphicx}
\usepackage{natbib}
\usepackage{caption}
\usepackage{dsfont}
\usepackage{colortbl}

\usepackage[utf8]{inputenc}
\usepackage{booktabs}
\usepackage{amsfonts}
\usepackage{nicefrac}
\usepackage{microtype}
\usepackage{xcolor}
\usepackage{amssymb}
\usepackage{multirow}
\usepackage{algorithm}
\usepackage{algpseudocode}
\usepackage{tabularray}

\usepackage{amsmath}
\usepackage{amsthm}
\title{AdaPilot: Towards Scene-Adaptive Policy Learning for Cross-Generator Text-to-Image Quality Optimization}

\author{%
  Wenjin Liu\textsuperscript{1,2}, Fayuan Ke\textsuperscript{2}, Yue Lu\textsuperscript{2},
  Zhe Cui\textsuperscript{2,$\dagger$}, Anh Tuan Luu\textsuperscript{1}, Haoran Luo\textsuperscript{1,$\dagger$}
}
\affiliations{%
  \textsuperscript{1}Nanyang Technological University, Singapore
  \qquad \textsuperscript{2}Hithink Research, China\\
  \texttt{wenjinliu23@outlook.com, haoran.luo@ntu.edu.sg}
}

\begin{document}

\maketitle
\begingroup
\renewcommand{\thefootnote}{\fnsymbol{footnote}}
\footnotetext[2]{Corresponding authors.}
\endgroup

\begin{abstract}
Existing methods for improving text-to-image generation quality have progressed from generator fine-tuning and prompt optimization to reinforcement learning with multi-turn visual feedback. However, existing strategies are deeply coupled with specific generators and tasks, and the learned capabilities are difficult to generalize into a universal quality optimization policy. Therefore, we propose \textbf{AdaPilot}, which learns a scene-adaptive, cross-generator transferable quality optimization policy by formulating multi-turn image generation as a Markov Decision Process (MDP) and optimizing it via end-to-end reinforcement learning. Specifically, AdaPilot decouples the policy from generator internals to enable cross-generator transfer, introduces scene-aware rewards that adaptively align quality assessment dimensions with task semantics, and employs process-level rewards to model the evolution trajectory of image quality. Experimental results show AdaPilot outperforms baselines in generation quality and generalization. Separate cross-generator evaluations further show that a single policy transfers zero-shot to unseen generators while maintaining positive average gains across all evaluated generators. Our project is available at \url{https://github.com/QwenQKing/Ada_pilot}.
  
\end{abstract}

\section{Introduction}

Text-to-image diffusion models such as FLUX~\cite{blackforestlabs2025flux2}, Qwen-Image~\cite{wu2025qwen}, LongCat~\cite{team2025longcat} and Ovis-Image~\cite{wang2025ovis} have advanced photorealistic synthesis, compositional generation, and text rendering. To further improve quality, methods have progressed from generator fine-tuning and prompt optimization to multi-turn visual-feedback reinforcement learning. Yet existing strategies remain generator- and scene-specific, limiting generalization across heterogeneous tasks. Learning a scene-adaptive, cross-generator transferable quality policy remains a major challenge~\cite{cai2025z,chen2026survey}.

\begin{figure}[!t]
  \centering
  \includegraphics[width=0.98\columnwidth]{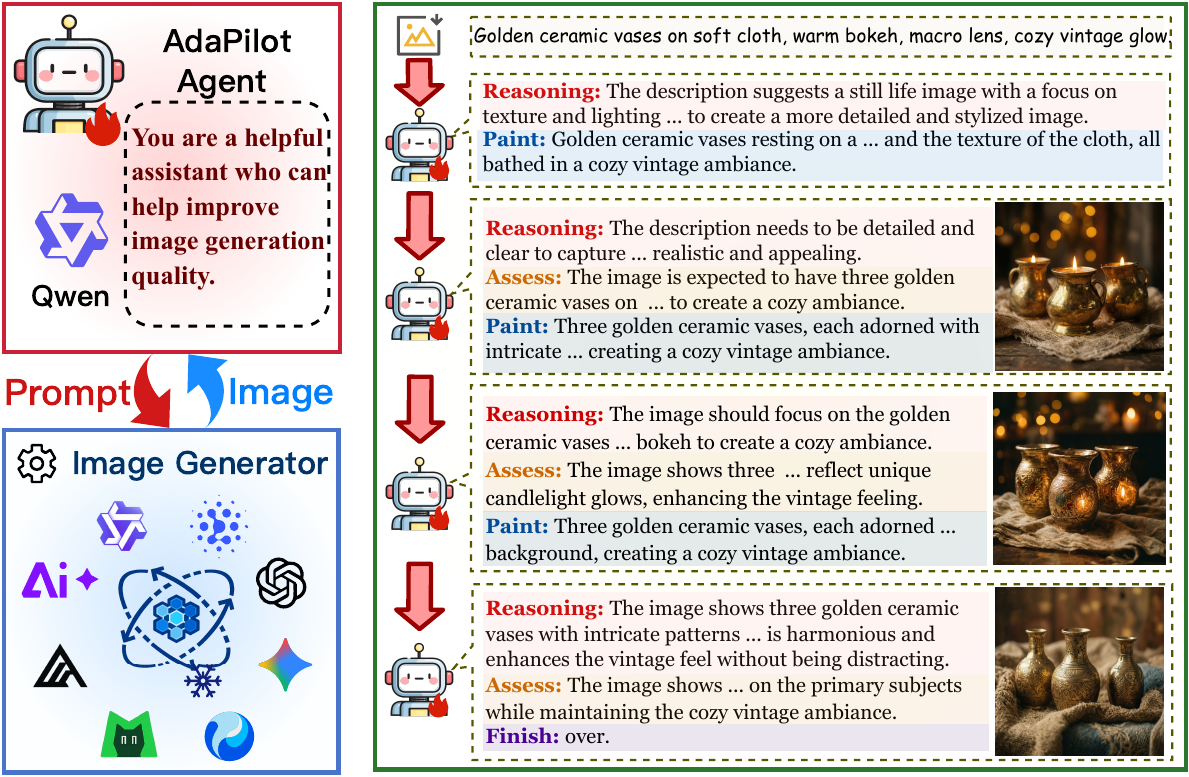}
  \caption{An illustration of the proposed AdaPilot.}
  \label{fig:fig-1}
\end{figure}

\begin{figure*}[t]
\centering
\includegraphics[width=\textwidth]{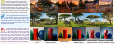}
\caption{Qualitative comparison with and without AdaPilot agent on Qwen-Image, GPT-Image-1, and Gemini-3-pro-Image-Preview. AdaPilot enhances the image generation quality across different generators without updating their parameters.}
\label{fig:results}
\end{figure*}

To optimize image generation quality, two categories of methods have been proposed~\cite{hartwig2025survey,cao2025controllable}. One updates generator parameters via generator-specific reinforcement learning (RL) or preference optimization, such as Flow-GRPO~\cite{liu2025flow}, Diffusion-DPO~\cite{wallace2024diffusion}, T2I-R1~\cite{jiang2025t2i}, and DanceGRPO~\cite{xue2025dancegrpo}. The other keeps the generator frozen, including prompt rewriting methods (Promptist~\cite{hao2023optimizing}, OPT2I~\cite{manas2024improving}, PromptEnhancer~\cite{wang2025promptenhancer}), iterative refinement methods (ReflectionFlow~\cite{zhuo2025reflection}, GenArtist~\cite{wang2024genartist}, Idea2Img~\cite{yang2024idea2img}), and multi-turn visual-feedback reinforcement learning agents~\cite{jiang2026genagent}.

However, existing methods still face three key challenges, including: \textit{(i)} \textbf{Cross-generator policy transfer remains limited.} Existing methods are typically designed or trained for specific generators and often require re-adaptation when the generator changes, making it difficult to develop reusable quality optimization capabilities. \textit{(ii)} \textbf{Reward signals lack scene adaptability.} Different scenes prioritize different quality dimensions, while fixed metric combinations struggle to provide accurate and targeted supervision across heterogeneous tasks. \textit{(iii)} \textbf{Multi-turn quality evolution lacks explicit modeling.} Existing methods primarily focus on final outputs, making it difficult to determine whether each refinement improves or degrades quality and thus to reliably guide round-by-round optimization throughout the interaction process.

To address these challenges, we propose AdaPilot (see Figure~\ref{fig:fig-1}), a scene-adaptive policy learning framework for cross-generator image quality optimization. AdaPilot formulates multi-turn generation as a visual-feedback MDP, enabling the policy to use intermediate results and adapt across scenes and generators. First, the policy relies solely on visually encoded observations, thereby supporting cross-generator transfer. Second, a scene-aware reward adaptively activates quality assessment dimensions aligned with task semantics and renormalizes their weights across different scenes. In addition, a process-level reward combines terminal image quality with a turn-wise monotonic improvement term, explicitly modeling quality evolution and discouraging quality regressions during multi-turn optimization. Training freezes the generator and optimizes only the policy end-to-end.

We evaluate AdaPilot on ten datasets across multiple frozen generators. Experimental results show that \textbf{AdaPilot} consistently outperforms the current baselines. These gains also extend to held-out quality dimensions excluded from the training reward. A policy trained on one generator drives unseen generators without adaptation while maintaining consistent gains across all evaluated generators (see Figure \ref{fig:results}), supporting robust transfer beyond the training generator.

\section{Related Work}
\label{related_work}
\textbf{Generator-Tuning Approaches.} Generator tuning with reward or preference signals~\cite{liu2026llm} couples objectives to model parameters and requires generator-specific re-optimization. Denoising Diffusion Policy Optimization~\cite{black2023training}, Direct Reward Fine-Tuning~\cite{clark2023directly}, AlignProp~\cite{prabhudesai2023aligning}, DPOK~\cite{fan2023dpok}, Temporal Diffusion Policy Optimization~\cite{zhang2024confronting}, Deep Reward Tuning~\cite{wu2024deep}, and Proximal Reward Difference Prediction~\cite{deng2024prdp} optimize diffusion models via policy gradients or reward backpropagation. ImageReward~\cite{xu2023imagereward} models human preferences; Diffusion-DPO~\cite{wallace2024diffusion}, Rich Human Feedback~\cite{liang2024rich}, Self-Play Fine-Tuning~\cite{yuan2024self}, and CoMat~\cite{jiang2024comat} align generators via preference or concept supervision. Flow-GRPO~\cite{liu2025flow} and DanceGRPO~\cite{xue2025dancegrpo} train generators with online RL, while T2I-R1~\cite{jiang2025t2i} adds collaborative semantic- and token-level CoT.

\textbf{Generator-Frozen Approaches.} Other approaches retain a pretrained generator and rely on external mechanisms. Promptist~\cite{hao2023optimizing} and PromptEnhancer~\cite{wang2025promptenhancer} optimize prompts through RL or CoT rewriting; OPT2I~\cite{manas2024improving} performs iterative LLM-based optimization. StructureDiffusion~\cite{feng2022training} and Attend-and-Excite~\cite{chefer2023attend} guide attention; MultiDiffusion~\cite{bar2023multidiffusion}, GLIGEN~\cite{li2023gligen}, and RealCompo~\cite{zhang2024realcompo} provide spatial control through diffusion composition or grounding; LayoutGPT~\cite{feng2023layoutgpt} and RPG~\cite{yang2024mastering} use LLM planning. Self-correcting LLM-controlled Diffusion~\cite{wu2024self}, Idea2Img~\cite{yang2024idea2img}, GenArtist~\cite{wang2024genartist}, and ReflectionFlow~\cite{zhuo2025reflection} refine outputs through verification, self-correction, or test-time scaling. GenAgent~\cite{jiang2026genagent} combines SFT and agentic RL with final-quality and reflection rewards for cross-tool generalization. AdaPilot instead couples scene-conditioned multi-metric reward routing with explicit turn-wise monotonicity supervision to stabilize quality across iterative turns.

\begin{figure*}[t]
\centering
\includegraphics[width=0.96\linewidth]{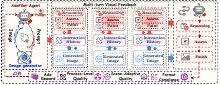}
\caption{Overview of AdaPilot, a policy for scene-adaptive image quality optimization through multi-turn visual feedback.}
\label{fig:AdaPilot}
\end{figure*}

\section{Methodology}
\label{sec:methodology}
We introduce AdaPilot (see Figure~\ref{fig:AdaPilot}), including adaptive multi-turn image generation, scene-aware quality-driven optimization, and cross-generator policy generalization.

\subsection{Adaptive Multi-Turn Image Generation}
\label{sec:interaction}

AdaPilot formulates multi-turn image generation as a Markov Decision Process, with a multimodal large language model (MLLM) $\mathcal{M}_\theta$ as the agent policy $\pi_\theta$ and a frozen image generator $\mathcal{G}_\phi$ as the environment $\mathcal{E}$. Given a description $q$, the agent observes intermediate images, diagnoses quality defects, and iteratively refines prompts through visual feedback.

\textbf{Environment and State.}
At interaction turn $t$, the state $s_t$ is the accumulated context and the action $a_t$ is the agent output. Whenever $a_t$ contains a generation instruction with optimized prompt $p_t$, the generator returns image $\mathcal{I}_t=\mathcal{G}_\phi(p_t)$. The image is encoded by the agent's visual encoder $\mathrm{Enc}_{\mathrm{vis}}$ and paired with a system-injected guiding prompt $u_t$ to form observation $o_t$; the resulting transition and state update are:
\begin{equation}
\begin{aligned}
\mathcal{I}_t &= \mathcal{G}_\phi(p_t),
&o_t &= \bigl(\mathrm{Enc}_{\mathrm{vis}}(\mathcal{I}_t),u_t\bigr),\\
s_1 &= q,
&s_{t+1} &= s_t\oplus a_t\oplus o_t,
\end{aligned}
\end{equation}
where $\oplus$ represents the sequence concatenation. 

\textbf{Structured Action Space.}
The agent is initialized with structured system prompts (see Figure~\ref{fig:agent-prompt}). Its action $a_t$ consists of reasoning reflection $a_t^{\mathrm{rea}}$ for identifying quality gaps, generation instruction $a_t^{\mathrm{gen}}$ for producing $p_t$, quality assessment $a_t^{\mathrm{ass}}$ for evaluating the returned image, and termination $a_t^{\mathrm{fin}}$ for returning the final result. At the first turn, the action likelihood factorizes over its structured sub-actions as:
\begin{equation}
\log \pi_\theta(a_1\mid s_1)
=\log \pi_\theta(a_1^{\mathrm{rea}}\mid s_1)
+\log \pi_\theta(a_1^{\mathrm{gen}}\mid s_1,a_1^{\mathrm{rea}}).
\end{equation}
For subsequent turns $t>1$, the agent reasons, assesses the latest image, and selects $a_t^{\mathrm{out}}\in\{a_t^{\mathrm{gen}},a_t^{\mathrm{fin}}\}$ to refine or terminate according to the current quality assessment. Let $h_t=(s_t,a_t^{\mathrm{rea}})$ denote the post-reasoning context:
\begin{equation}
\begin{aligned}
\log \pi_\theta(a_t \mid s_t)
={}& \log \pi_\theta(a_t^{\text{rea}} \mid s_t) \\
&\hspace{-3.2em}+ \log \pi_\theta(a_t^{\text{ass}}\mid h_t)
+ \log \pi_\theta(a_t^{\text{out}}\mid h_t,a_t^{\text{ass}}).
\end{aligned}
\end{equation}
\textbf{Generation Trajectory.}
A complete trajectory $\tau$ records at most $T_{\max}$ turns. We set $\tilde o_t=o_t$ after generation and $\tilde o_t=\varnothing$ otherwise. The final image $\mathcal{I}^*$ is returned by the last generation instruction:
\begin{equation}
\begin{split}
\tau &= \bigl((s_1,a_1,\tilde o_1),\ldots,(s_T,a_T,\tilde o_T)\bigr),\\
t^* &= \max\{t\leq T:a_t^{\mathrm{gen}}\in a_t\},
\qquad \mathcal{I}^*=\mathcal{I}_{t^*}.
\end{split}
\end{equation}
A trajectory terminates upon $a_T^{\mathrm{fin}}$ or at the turn budget $T_{\max}$.

\par\noindent\textbf{Proposition 1.} \textit{Multi-turn visual feedback can achieve higher final image quality than single-turn refinement through iterative defect correction.}
\begin{proof}
We provide experimental results in Section~\ref{sec:exp_turn} and theoretical proofs in Appendix~\ref{app:proof1}.
\end{proof}

\definecolor{tagreasoning}{rgb}{0.90,0.45,0.05}
\definecolor{tagpaint}{rgb}{0.85,0.18,0.10}
\definecolor{tagassess}{rgb}{0.80,0.55,0.00}
\definecolor{tagfinish}{rgb}{0.78,0.10,0.35}
\definecolor{description}{rgb}{0.1,0.1,0.75}
\begin{figure*}[t]
\centering
\includegraphics[width=1\textwidth]{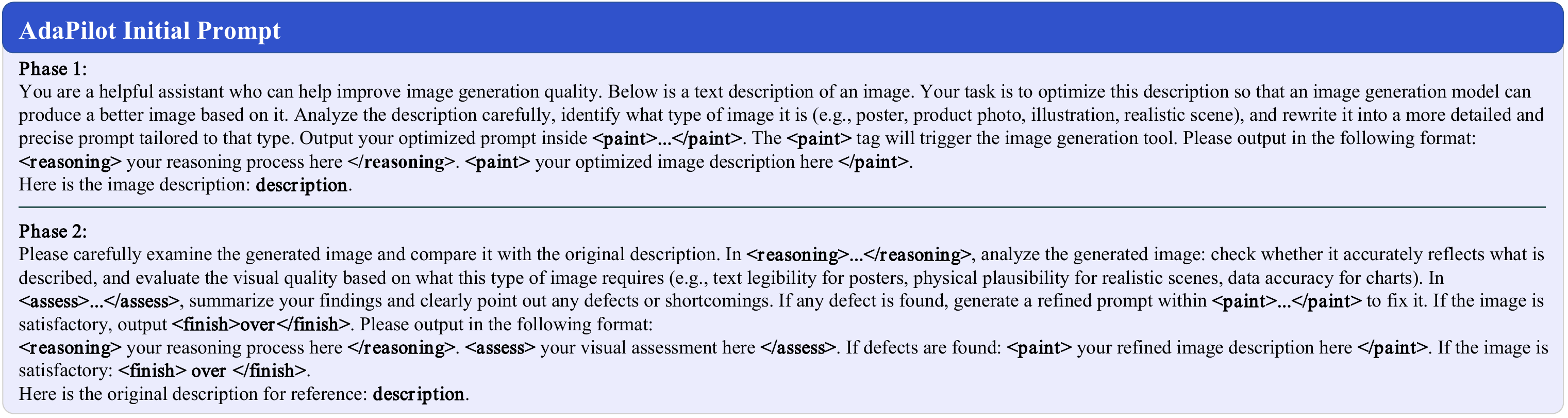}
\caption{The system prompts for the AdaPilot agent. The agent optimizes the text prompt in Phase~1 and assesses the generated image against the original description, iteratively refining the prompt in Phase~2 until the scene-specific criteria are fully met.}
\label{fig:agent-prompt}
\end{figure*}

\subsection{Scene-Aware Quality-Driven Optimization}
\label{sec:reward}

AdaPilot introduces \textbf{AdaReward} to supervise turn-wise quality evolution through scene-adaptive assessment and process-level improvement modeling within each trajectory.

\textbf{Overall Reward $R(\tau)$.}
For a trajectory $\tau$ containing $n$ generated images, let $R_{\mathrm{fmt}}(\tau)$ be its format-compliance score, $Q_j$ the quality of its $j$-th image for $j\in\{1,\ldots,n\}$, and $\mathrm{mono}(\tau)$ its monotonic-improvement score. AdaReward gates quality supervision by interaction-format compliance:
\begin{equation}
R(\tau)=
\begin{cases}
-1+R_{\mathrm{fmt}}(\tau), & R_{\mathrm{fmt}}<1\ \lor\ n=0,\\
\lambda_Q Q_n+\lambda_M\mathrm{mono}(\tau), & R_{\mathrm{fmt}}=1,\ n>0,
\end{cases}
\end{equation}
where $\lambda_Q,\lambda_M\geq0$ and $\lambda_Q+\lambda_M=1$ balance terminal quality and process-level improvement, respectively.

\textbf{Format Compliance $R_{\text{fmt}}$.}
Each sub-action uses a structured tag: \texttt{<reasoning>}, \texttt{<paint>}, \texttt{<assess>}, or \texttt{<finish>}. Define $C_t=\mathbf{1}[\text{round }t\text{ uses its required tags}]$, where $\mathbf{1}[\cdot]$ is the indicator function. The role-weighted compliance score is defined over the complete trajectory as:
\begin{equation}
R_{\mathrm{fmt}}(\tau)=
\min\!\left(\sum_{t=1}^{T}w_{k(t)}C_t,1\right),
\end{equation}
where $k(t)$ identifies turn $t$ as first, intermediate, or last, and the weights satisfy $w_{k(t)}>0$ and $\sum_{t=1}^{T}w_{k(t)}=1$. Thus, $R_{\mathrm{fmt}}=1$ only for fully compliant trajectories, disabling quality supervision otherwise by construction.

\textbf{Scene-Adaptive Quality Assessment.}
Let $\mathcal{K}(q)$ be the metric subset activated for the scene described by $q$, and let $r_k(\mathcal{I}_j,q)\in[0,1]$ be the normalized score of metric $k$. The scene-adaptive quality score for image $\mathcal{I}_j$ is computed as follows:
\begin{equation}
Q_j=\sum_{k\in\mathcal{K}(q)}\bar w_k r_k(\mathcal{I}_j,q),
\qquad
\bar w_k=\frac{w_k}{\sum_{k'\in\mathcal{K}(q)}w_{k'}},
\end{equation}
where $w_k$ is the base metric weight and $\sum_{k\in\mathcal{K}(q)}\bar w_k=1$. The available metrics assess text-image alignment, human preference, aesthetics, object accuracy, and text rendering through CLIP-T, HPS, Aesthetic, Grounding DINO, and OCR scores, respectively. Scene-conditioned activation and weight renormalization align supervision with task-specific quality requirements across heterogeneous generation tasks.

\textbf{Process-Level Quality Modeling.}
AdaReward explicitly rewards trajectories whose image quality improves at every generation step rather than only at termination:
\begin{equation}
\mathrm{mono}(\tau)=
\begin{cases}
\displaystyle\frac{Q_n-Q_1}{\max(1-Q_1,\epsilon_0)},
& \substack{n\geq2,\\ Q_1<Q_2<\cdots<Q_n},\\
0, & \text{otherwise},
\end{cases}
\end{equation}
where $\epsilon_0>0$ is a numerical-stability constant. Any quality regression sets this term to zero, while a fully improving trajectory is rewarded by its normalized gain from $Q_1$ to $Q_n$.

\textbf{End-to-End Policy Optimization.}
AdaPilot adopts GRPO for end-to-end policy optimization. For each $q\sim\mathcal{D}$, it samples $N$ trajectories $\{\tau_i\}_{i=1}^{N}$ from the old policy $\pi_{\theta_{\mathrm{old}}}$. With $R_i=R(\tau_i)$, the group statistics and normalized advantage are jointly estimated from all trajectory rewards in the group:
\begin{equation}
\mu_q=\frac{1}{N}\sum_{i=1}^{N}R_i,\;
\sigma_q^2=\frac{1}{N}\sum_{i=1}^{N}(R_i-\mu_q)^2,\;
\hat A_i=\frac{R_i-\mu_q}{\sigma_q+\epsilon_0}.
\end{equation}
For trajectory $\tau_i$, $\hat A_i$ is shared by all token positions $\ell$ in the agent's autoregressive set $\mathcal{T}_m^{(i)}$ selected by mask $m$. Let $s_{i,\ell}$ and $a_{i,\ell}$ be the state and generated token at position $\ell$. The per-token importance ratio $\rho_{i,\ell}$ and its clipped counterpart $\bar\rho_{i,\ell}$ are defined as:
\begin{equation}
\begin{aligned}
\rho_{i,\ell}
&=\frac{\pi_\theta(a_{i,\ell}\mid s_{i,\ell})}
{\pi_{\theta_{\mathrm{old}}}(a_{i,\ell}\mid s_{i,\ell})},\\
\bar\rho_{i,\ell}
&=\operatorname{clip}
(\rho_{i,\ell},1-\epsilon_c,1+\epsilon_c),
\end{aligned}
\end{equation}
where $\epsilon_c>0$ is the clipping range; the training loss is:
\begin{equation}
\begin{aligned}
\mathcal{L}_{\mathrm{train}}(\theta)
={}&-\mathbb{E}
\Bigg[
\frac{1}{N}\sum_{i=1}^{N}
\frac{1}{|\mathcal{T}_m^{(i)}|}
\\[-2pt]
&\quad\sum_{\ell\in\mathcal{T}_m^{(i)}}\Bigl[
\min\left(
\rho_{i,\ell}\hat A_i,
\bar\rho_{i,\ell}\hat A_i
\right)
-\beta_{\mathrm{KL}}D_{i,\ell}^{\mathrm{KL}}
\Bigr]
\Bigg].
\end{aligned}
\end{equation}
where $D_{i,\ell}^{\mathrm{KL}}=D_{\mathrm{KL}}(\pi_\theta\|\pi_{\theta_0})$ is the state-wise KL to the fixed anchor, weighted by $\beta_{\mathrm{KL}}\geq0$. Expectations follow the trajectory sampling above; gradients update only $\theta$ on $\mathcal{T}_m^{(i)}$, while $\phi$ remains frozen throughout end-to-end training.

\par\noindent\textbf{Proposition 2.} \textit{Scene-adaptive multi-metric supervision can enhance quality optimization across heterogeneous tasks by matching rewards to scene-specific requirements.}
\begin{proof}
We provide experimental results in Section~\ref{sec:exp_ablation} and theoretical proofs in Appendix~\ref{app:proof2}.
\end{proof}

\begin{table*}[!t]
\centering
\begingroup
\fontsize{9}{10}\selectfont
\begin{tblr}{
  width=\textwidth,
  colspec={
    Q[l,1.9cm]
    Q[l,1.35cm]
    X[c,1.08]X[c]X[c]X[c]X[c]
    X[c,1.25]X[c]X[c]X[c,1.25]X[c,2.1]
  },
  colsep=0.6mm,
  rowsep=0.6pt,
  stretch=1.0,
  row{1}={font=\bfseries,halign=c,valign=m},
  cell{1}{1}={halign=l},
  cell{1}{2}={halign=l},
  cell{2}{1}={r=4}{},
  cell{6}{1}={r=5}{},
  cell{11}{1}={r=4}{},
  cell{15}{1}={r=3}{},
  cell{18}{1}={r=4}{},
  cell{22}{1}={r=4}{},
  cell{26}{1}={r=7}{},
  hline{1,Z}={1pt},
  hline{2}={0.6pt},
  hline{6,11,15,18,22,26}={0.3pt}
}
\textbf{Dataset} & \textbf{Metric} & \textbf{Baseline} & \textbf{CoT} & \textbf{GRPO} & \textbf{SFT} & \textbf{BoN} & \mbox{\textbf{F-GRPO}} & \textbf{T2I-R1} & \textbf{Copilot} & \textbf{MPrompt} & \mbox{\textbf{Ours ($\Delta\uparrow$)}} \\
\textbf{Fine-T2I} & GDino & 80.63 & 74.47 & 74.41 & 73.58 & 80.15 & 77.95 & 80.57 & 75.86 & 73.60 & \textbf{84.25}\,\textbf{(+3.62)} \\
 & Realism & 41.47 & 40.43 & 40.04 & 38.48 & 42.58 & 30.43 & 24.41 & 41.41 & 40.25 & \textbf{46.46}\,\textbf{(+4.99)} \\
 & HPS & 71.38 & 74.85 & 71.62 & 76.59 & 71.57 & 17.26 & 68.48 & 70.08 & 66.38 & \textbf{77.90}\,\textbf{(+6.52)} \\
 & Physics & 38.10 & 35.63 & 35.24 & 35.16 & 36.42 & 26.84 & 26.56 & 37.10 & 30.17 & \textbf{45.83}\,\textbf{(+7.73)} \\
\textbf{AutoPP1M} & CLIP-T & 95.84 & 96.74 & 96.98 & 96.27 & 97.21 & 87.54 & 97.06 & 94.37 & 96.74 & \textbf{97.31}\,\textbf{(+1.47)} \\
 & GDino & 90.39 & 87.21 & 84.11 & 85.62 & 86.73 & 82.03 & 83.74 & 88.17 & 87.47 & \textbf{91.75}\,\textbf{(+1.36)} \\
 & Realism & 43.95 & 48.44 & 45.70 & 46.24 & 43.36 & 27.54 & 27.54 & 45.85 & 36.44 & \textbf{54.49}\,\textbf{(+10.54)} \\
 & HPS & 52.04 & 54.39 & 51.98 & 56.68 & 51.61 & 14.82 & 55.43 & 50.61 & 50.93 & \textbf{57.37}\,\textbf{(+5.33)} \\
 & Physics & 45.51 & 47.07 & 44.34 & 43.37 & 44.73 & 25.36 & 29.88 & 46.74 & 23.73 & \textbf{59.77}\,\textbf{(+14.26)} \\
\textbf{HQ-Poster} & Aesthetic & 56.64 & 56.64 & 50.20 & 53.32 & 58.98 & 57.69 & 24.22 & 61.25 & 64.68 & \textbf{77.15}\,\textbf{(+20.51)} \\
 & Realism & 39.02 & 38.48 & 34.18 & 35.55 & 39.65 & 20.38 & 17.19 & 36.43 & 28.17 & \textbf{49.41}\,\textbf{(+10.39)} \\
 & HPS & 65.81 & 66.68 & 66.54 & 67.23 & 65.02 & 10.04 & 43.07 & 57.38 & 64.13 & \textbf{68.65}\,\textbf{(+2.84)} \\
 & Physics & 41.74 & 40.16 & 33.13 & 34.25 & 41.27 & 21.15 & 25.79 & 41.03 & 43.95 & \textbf{54.13}\,\textbf{(+12.39)} \\
\textbf{Lunara} & CLIP-T & 84.99 & 95.84 & 95.48 & 97.44 & 96.69 & 90.88 & 94.43 & 90.16 & 94.87 & \textbf{97.71}\,\textbf{(+12.72)} \\
 & Aesthetic & 89.77 & 86.52 & 87.50 & 90.62 & 88.09 & 78.71 & 66.80 & 86.28 & 88.41 & \textbf{92.77}\,\textbf{(+3.00)} \\
 & HPS & 74.38 & 69.13 & 69.12 & 72.85 & 70.68 & 14.39 & 75.29 & 71.20 & 72.41 & \textbf{76.54}\,\textbf{(+2.16)} \\
\textbf{FinMME} & CLIP-T & 94.54 & 92.34 & 93.32 & 87.57 & 93.11 & 46.45 & 3.15 & 90.52 & 61.75 & \textbf{96.36}\,\textbf{(+1.82)} \\
 & OCR & 5.36 & 21.37 & 11.20 & 15.83 & 2.33 & 0.87 & 2.01 & 4.64 & 4.28 & \textbf{22.76}\,\textbf{(+17.40)} \\
 & Realism & 5.13 & 7.42 & 6.64 & 5.66 & 6.05 & 2.17 & 7.62 & 7.03 & 3.57 & \textbf{13.28}\,\textbf{(+8.15)} \\
 & HPS & 12.28 & 15.65 & 15.98 & 14.78 & 15.10 & 0.06 & 0.13 & 8.93 & 13.91 & \textbf{16.63}\,\textbf{(+4.35)} \\
\textbf{TextAtlas5M} & CLIP-T & 95.30 & 92.34 & 93.18 & 95.34 & 94.52 & 75.56 & 94.27 & 92.27 & 89.82 & \textbf{95.97}\,\textbf{(+0.67)} \\
 & OCR & 33.03 & 22.03 & 52.17 & 47.38 & 28.47 & 7.24 & 3.59 & 40.37 & 9.38 & \textbf{58.35}\,\textbf{(+25.32)} \\
 & Aesthetic & 31.64 & 41.41 & 36.33 & 36.52 & 31.64 & 30.56 & 11.33 & 32.24 & 28.85 & \textbf{48.63}\,\textbf{(+16.99)} \\
 & HPS & 46.18 & 48.08 & 46.11 & 49.70 & 43.79 & 3.54 & 28.06 & 48.96 & 28.51 & \textbf{52.13}\,\textbf{(+5.95)} \\
\textbf{AVG} & CLIP-T & 92.67 & 94.31 & 94.74 & 94.16 & 95.38 & 75.11 & 72.23 & 91.83 & 85.80 & \textbf{96.84}\,\textbf{(+4.17)} \\
 & GDino & 85.51 & 80.84 & 79.26 & 79.60 & 83.44 & 79.99 & 82.16 & 82.02 & 80.53 & \textbf{88.00}\,\textbf{(+2.49)} \\
 & OCR & 19.19 & 21.70 & 31.69 & 31.61 & 15.40 & 4.06 & 2.80 & 22.50 & 6.83 & \textbf{40.55}\,\textbf{(+21.36)} \\
 & Aesthetic & 59.35 & 61.52 & 58.01 & 60.15 & 59.57 & 55.65 & 34.12 & 59.92 & 60.65 & \textbf{72.85}\,\textbf{(+13.50)} \\
 & Realism & 32.39 & 33.69 & 31.64 & 31.48 & 32.91 & 20.13 & 19.19 & 32.68 & 27.11 & \textbf{40.91}\,\textbf{(+8.52)} \\
 & HPS & 53.68 & 54.80 & 53.56 & 56.30 & 52.96 & 10.02 & 45.08 & 51.19 & 49.38 & \textbf{58.20}\,\textbf{(+4.53)} \\
 & Physics & 41.78 & 40.95 & 37.57 & 37.59 & 40.81 & 24.45 & 27.41 & 41.62 & 32.62 & \textbf{53.24}\,\textbf{(+11.46)} \\
\end{tblr}
\endgroup
\caption{Comparison of the baselines and AdaPilot on six in-distribution datasets covering diverse tasks and domains. F-GRPO denotes Flow-GRPO, and Copilot denotes T2I-Copilot. $\Delta\uparrow$ means the gap between AdaPilot and the baseline image generator (Qwen-Image), where higher values indicate better performance. Bold values mean the best performance. All values are in \%.}
\label{tab:main-id}
\end{table*}

\begin{table*}[t]
\centering
\begingroup
\fontsize{9}{10}\selectfont
\begin{tblr}{
  width=\textwidth,
  colspec={
    Q[l,1.9cm]
    Q[l,1.35cm]
    X[c,1.08]X[c]X[c]X[c]X[c]
    X[c,1.25]X[c]X[c]X[c,1.25]X[c,2.1]
  },
  colsep=0.6mm,
  rowsep=0.6pt,
  stretch=1.0,
  row{1}={font=\bfseries,halign=c,valign=m},
  cell{1}{1}={halign=l},
  cell{1}{2}={halign=l},
  cell{2}{1}={r=4}{},
  cell{6}{1}={r=3}{},
  cell{9}{1}={r=3}{},
  cell{12}{1}={r=3}{},
  hline{1,Z}={1pt},
  hline{2}={0.6pt},
  hline{6,9,12}={0.3pt}
}
\textbf{Dataset} & \textbf{Metric} & \textbf{Baseline} & \textbf{CoT} & \textbf{GRPO} & \textbf{SFT} & \textbf{BoN} & \mbox{\textbf{F-GRPO}} & \textbf{T2I-R1} & \textbf{Copilot} & \textbf{MPrompt} & \mbox{\textbf{Ours ($\Delta\uparrow$)}} \\
\textbf{DPG-Bench} & CLIP-T & 94.89 & 94.43 & 92.92 & 94.92 & 94.54 & 81.93 & 92.95 & 95.24 & 94.71 & \textbf{95.72}\,\textbf{(+0.83)} \\
 & GDino & 82.68 & 79.42 & 76.11 & 78.37 & 79.46 & 67.47 & 77.42 & 84.35 & 83.10 & \textbf{86.67}\,\textbf{(+3.99)} \\
 & Realism & 44.40 & 45.51 & 41.60 & 44.34 & 44.73 & 28.08 & 26.76 & 42.38 & 36.46 & \textbf{52.36}\,\textbf{(+7.96)} \\
 & Physics & 42.21 & 41.94 & 37.40 & 38.29 & 41.13 & 23.81 & 26.40 & 43.76 & 22.89 & \textbf{51.40}\,\textbf{(+9.19)} \\
\textbf{TVBench} & OCR & 39.14 & 35.19 & 42.14 & 43.32 & 39.87 & 4.61 & 0.09 & 37.76 & 42.64 & \textbf{44.60}\,\textbf{(+5.46)} \\
 & Aesthetic & 15.82 & 6.25 & 8.98 & 6.25 & 13.67 & 9.38 & 5.47 & 12.65 & 7.59 & \textbf{24.02}\,\textbf{(+8.20)} \\
 & HPS & 14.31 & 20.72 & 15.60 & 14.29 & 15.96 & 0.17 & 0.32 & 15.02 & 0.53 & \textbf{21.81}\,\textbf{(+7.50)} \\
\textbf{CompBench} & GDino & 94.53 & 84.74 & 87.34 & 88.93 & 91.80 & 71.54 & 86.62 & 89.74 & 93.38 & \textbf{96.78}\,\textbf{(+2.25)} \\
 & Realism & 51.17 & 45.12 & 45.51 & 47.27 & 48.05 & 32.71 & 25.39 & 52.31 & 47.79 & \textbf{55.86}\,\textbf{(+4.69)} \\
 & Physics & 34.96 & 40.39 & 35.16 & 38.09 & 30.66 & 18.21 & 17.77 & 35.94 & 28.68 & \textbf{45.90}\,\textbf{(+10.94)} \\
\textbf{TIIF-Bench} & GDino & 89.53 & 77.53 & 81.64 & 83.76 & 86.58 & 77.30 & 90.57 & 85.64 & 84.39 & \textbf{93.47}\,\textbf{(+3.94)} \\
 & Realism & 43.95 & 47.22 & 40.43 & 42.58 & 44.14 & 31.45 & 22.64 & 40.17 & 39.58 & \textbf{53.71}\,\textbf{(+9.76)} \\
 & Physics & 35.69 & 38.09 & 33.73 & 37.89 & 31.94 & 28.54 & 19.60 & 34.76 & 25.00 & \textbf{50.39}\,\textbf{(+14.70)} \\
\end{tblr}
\endgroup
\caption{Comparison of the baselines and the proposed AdaPilot on four out-of-distribution datasets. $\Delta\uparrow$ is AdaPilot's gain over Qwen-Image, and bold marks the best result. All values are in \%.}
\label{tab:main-ood}
\end{table*}

\begin{figure*}[t]
\centering
\includegraphics[width=\textwidth]{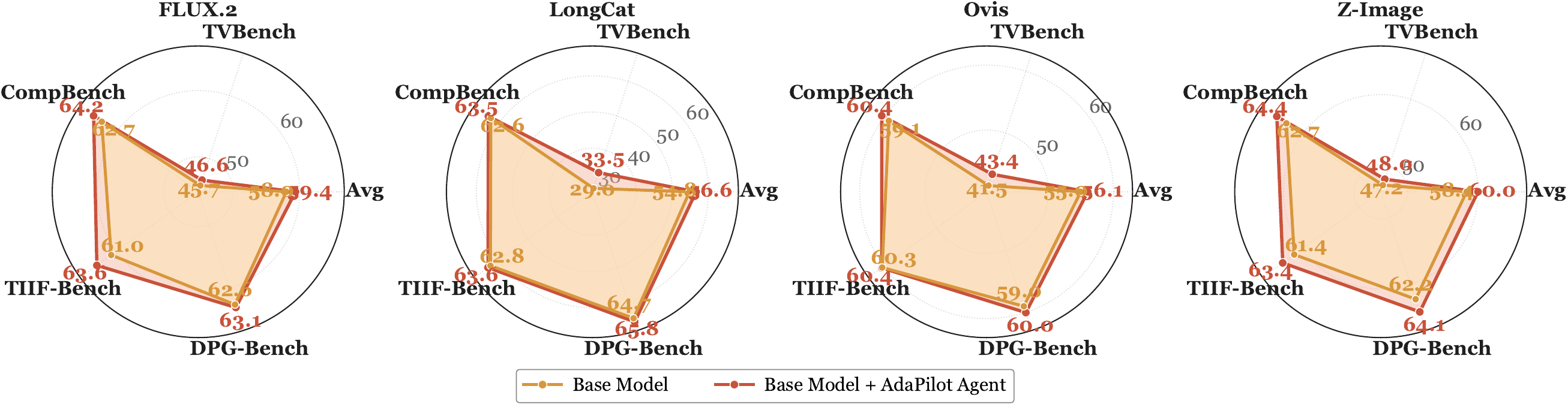}
\caption{Averages over applicable evaluation metrics for four unseen image generators on four out-of-distribution datasets and overall, comparing results with and without AdaPilot to evaluate zero-shot cross-generator transfer.}
\label{fig:ood_radar}
\end{figure*}

\begin{table*}[t]
\centering
\begingroup
\fontsize{9}{10}\selectfont
\begin{tblr}{
  width=\textwidth,
  colspec={
    Q[l,4.0cm]
    X[c]X[c]X[c]X[c]
    X[c]X[c]X[c]X[c]
  },
  colsep=0.6mm,
  rowsep=0.6pt,
  stretch=1.0,
  row{1-2}={font=\bfseries,halign=c,valign=m},
  cell{1}{1}={r=2}{halign=l,valign=m},
  cell{1}{2}={c=2}{halign=c},
  cell{1}{4}={c=2}{halign=c},
  cell{1}{6}={c=2}{halign=c},
  cell{1}{8}={c=2}{halign=c},
  hline{1,Z}={1pt},
  hline{2}={2-3,4-5,6-7,8-9}{0.5pt},
  hline{3}={0.6pt},
  hline{7}={0.6pt}
}
\textbf{Method} & \textbf{TextAtlas5M} & & \textbf{AutoPP1M} & & \textbf{DPG-Bench} & & \textbf{CompBench} & \\
 & \textbf{OCR$\uparrow$} & \textbf{Aesthetic$\uparrow$} & \textbf{HPS$\uparrow$} & \textbf{Realism$\uparrow$} & \textbf{HPS$\uparrow$} & \textbf{Realism$\uparrow$} & \textbf{HPS$\uparrow$} & \textbf{CLIP-T$\uparrow$} \\
\textit{w/o} Agent & 33.03 & 31.64 & 52.04 & 43.95 & 55.98 & 44.40 & 65.90 & 82.50 \\
\textit{w/o} RL & 23.09 & 30.17 & 55.15 & 40.45 & 60.72 & 39.14 & 68.31 & 82.59 \\
\textit{w/o} Scene-Adaptive Reward & 48.30 & 37.89 & 52.83 & 46.88 & 57.86 & 45.90 & 67.81 & 85.78 \\
\textit{w/o} Monotonic Reward & 34.30 & 35.24 & 55.90 & 48.24 & 58.23 & 47.63 & 70.06 & 84.19 \\
\textbf{AdaPilot (Full)} & \textbf{58.35} & \textbf{48.63} & \textbf{57.37} & \textbf{54.49} & \textbf{60.77} & \textbf{52.36} & \textbf{72.06} & \textbf{86.67} \\
\end{tblr}
\endgroup
\caption{
Ablation study.
\textit{w/o} RL is without reinforcement learning;
\textit{w/o} Agent is without the AdaPilot Agent;
\textit{w/o} Scene-Adaptive Reward is without Scene-Adaptive Quality Assessment;
\textit{w/o} Monotonic Reward is without the Monotonic Improvement term.
}
\label{tab:ablation}
\end{table*}

\subsection{Cross-Generator Policy Generalization}
\label{sec:transfer}

AdaPilot accesses generators only through prompts and encoded visual observations, never their internals or gradients. This shared interaction space decouples the policy from generators and enables cross-generator reuse without retraining.

\textbf{Training.}
Let $\mathcal{G}_{\phi}^{\mathrm{train}}$ denote the frozen generator used for training and $\mathcal{D}$ the training-description distribution. AdaPilot learns the policy parameters $\theta^*$ by optimizing rewards over trajectories produced through interaction with this generator:
\begin{equation}
\theta^*
=
\arg\max_{\theta}
\mathbb{E}_{q\sim\mathcal{D}}
\mathbb{E}_{\tau\sim
\pi_\theta(\cdot\mid q;\mathcal{G}_{\phi}^{\mathrm{train}})}
\left[R(\tau)\right],
\qquad
\phi\ \text{fixed}.
\end{equation}
Here the semicolon indicates that $\mathcal{G}_{\phi}^{\mathrm{train}}$ supplies environment transitions but is not part of the optimized policy. Thus, reward-driven updates are confined to $\theta$ throughout training.

\textbf{Inference.}
At inference, an unseen generator $\mathcal{G}_{\phi}^{\mathrm{infer}}\neq\mathcal{G}_{\phi}^{\mathrm{train}}$ replaces the training generator, while the optimized policy is reused without adaptation. The resulting observation and reused policy parameters are therefore given by:
\begin{equation}
o_t^{\text{infer}}
=\bigl(
\mathrm{Enc}_{\mathrm{vis}}
(\mathcal{G}_{\phi}^{\mathrm{infer}}(p_t)),
u_t
\bigr),
\qquad
\theta^{\mathrm{infer}}=\theta^*.
\end{equation}
The visual encoder maps outputs from different generators into the same observation space, allowing $\pi_{\theta^*}$ to apply its learned reasoning, assessment, and refinement behavior without updating either the policy or the new generator.

\par\noindent\textbf{Proposition 3.} \textit{Policy-generator decoupling can enable a single policy to improve outputs from unseen generators zero-shot, without updating either model during transfer.}
\begin{proof}
We provide experimental results in Section~\ref{sec:exp_transfer} and theoretical proofs in Appendix~\ref{app:proof3}.
\end{proof}

\section{Experiments}
\label{sec:experiments}

This section reports the experimental setup, results, and analysis. We answer the following research questions (RQ).
\textbf{RQ1:} Does AdaPilot outperform existing baselines in generation quality?
\textbf{RQ2:} How well does AdaPilot generalize to out-of-distribution (OOD) datasets?
\textbf{RQ3:} Can a single policy transfer zero-shot to unseen generators?
\textbf{RQ4:} How do the individual components contribute to performance?
\textbf{RQ5:} Does image quality improve monotonically across turns?

\subsection{Experimental Setup}
\label{sec:exp_setup}

\textbf{Datasets.} We evaluate \textbf{AdaPilot} on ten datasets, including \textbf{Fine-T2I}~\cite{ma2026fine}, \textbf{DPG-Bench}~\cite{hu2024ella}, \textbf{T2I-CompBench}~\cite{huang2023t2i} (CompBench), \textbf{TIIF-Bench}~\cite{wei2025tiif}, \textbf{HQ-Poster-100K}~\cite{chen2025postercraft} (HQ-Poster), \textbf{TextAtlas5M}~\cite{wang2025textatlas5m}, \textbf{TableVisBench}~\cite{liu2025showtable} (TVBench), \textbf{AutoPP1M}~\cite{fan2026autopp}, \textbf{FinMME}~\cite{luo2025finmme} (using image-derived text descriptions), and \textbf{Lunara}~\cite{wang2026moonworkslunaraaestheticdataset}. More details are in Appendix~\ref{app:datasets}.

\textbf{Baselines.}
AdaPilot is compared with nine baselines:
\textbf{Baseline} (Qwen-Image~\cite{wu2025qwen}),
\textbf{CoT}~\cite{wei2022chain},
\textbf{GRPO}~\cite{guo2025deepseek},
Supervised Fine-Tuning (\textbf{SFT}),
Best-of-N (\textbf{BoN})~\cite{ichihara2025evaluation},
\textbf{Flow-GRPO}~\cite{liu2025flow},
\textbf{T2I-R1}~\cite{jiang2025t2i},
\textbf{T2I-Copilot}~\cite{chen2025t2i}, and
\textbf{MinorityPrompt} (\textbf{MPrompt})~\cite{um2024minority}.
More details are illustrated in Appendix~\ref{app:baselines}.

\textbf{Evaluation Metrics.} We evaluate \textbf{AdaPilot} with 7 metrics: \textbf{CLIP-T}, \textbf{HPS}, \textbf{Aesthetic}, \textbf{GDino}, \textbf{OCR}, \textbf{Realism}, and \textbf{Physics}. More details are illustrated in Appendix~\ref{app:metrics}.

\textbf{Implementation Details.}
We adopt Qwen2.5-VL-7B-Instruct (\textbf{Qwen2.5-VL-7B})~\cite{bai2025qwen25vltechnicalreport} as the agent $\mathcal{M}_\theta$, trained with Qwen-Image-2512 serving as the generator $\mathcal{G}_\phi$. All experiments run on two servers, each with eight NVIDIA H100 GPUs (80GB). More details are in Appendix~\ref{app:implementation}.

\definecolor{ds1}{rgb}{1.0000,0.9725,0.9020}
\definecolor{ds2}{rgb}{0.9961,0.9020,0.9059}
\definecolor{ds3}{rgb}{0.8980,0.9333,0.9686}
\definecolor{ds4}{rgb}{0.9255,0.8941,0.9608}
\definecolor{ds5}{rgb}{1.0000,0.9412,0.8588}
\definecolor{ds6}{rgb}{0.9882,0.9176,0.9098}
\definecolor{avg}{rgb}{0.9961,0.8627,0.8588}
\definecolor{dgreen}{rgb}{0.13,0.55,0.13}

\subsection{Main Results (RQ1)}
\label{sec:exp_main}

As shown in Table~\ref{tab:main-id}, AdaPilot outperforms all baselines across every evaluation dimension on datasets spanning fine-grained, e-commerce, poster, aesthetic, text-rich, and financial-chart scenes. Further results show:
\textit{(i)} \textbf{Generator-tuning baselines vary across tasks.} Flow-GRPO and T2I-R1 trail the Qwen-Image baseline on most reported metrics, while AdaPilot learns reusable policy-side optimization with a frozen generator.
\textit{(ii)} \textbf{Strongest gains align with targeted dimensions.} The largest improvements occur in text rendering and aesthetics, consistent with the quality dimensions selected by the scene-adaptive reward.
\textit{(iii)} \textbf{Policy learning works across optimization paradigms.} AdaPilot outperforms single-turn CoT, SFT, GRPO, and BoN, iterative MinorityPrompt, and multi-turn T2I-Copilot, supporting visual-feedback policy learning across diverse alternatives.

\begin{figure}[t]
  \centering
  \includegraphics[width=0.98\linewidth]{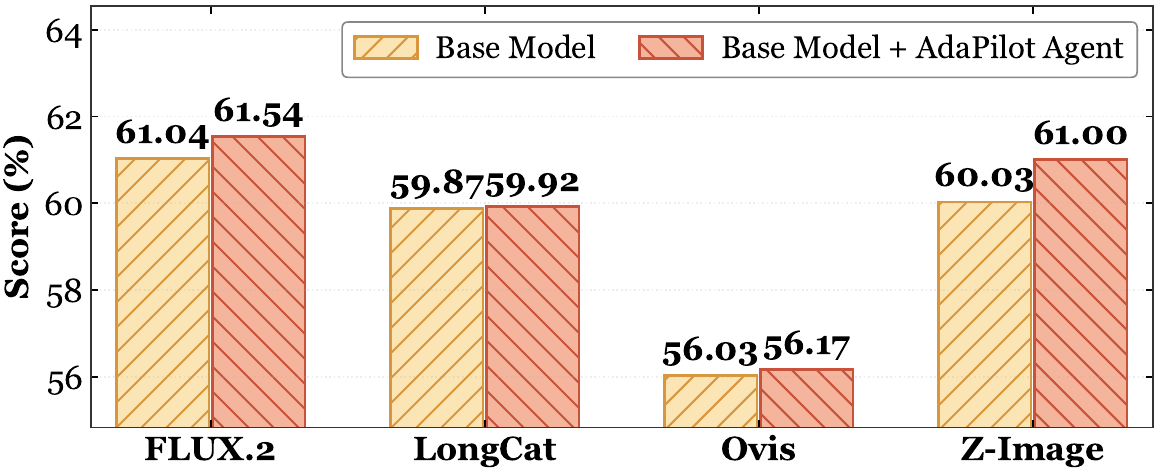}
  \caption{Average values of unseen generators on in-distribution datasets with and without AdaPilot agent.}
  \label{fig:id_avg}
\end{figure}

\begin{figure*}[t]
\centering
\includegraphics[width=\textwidth]{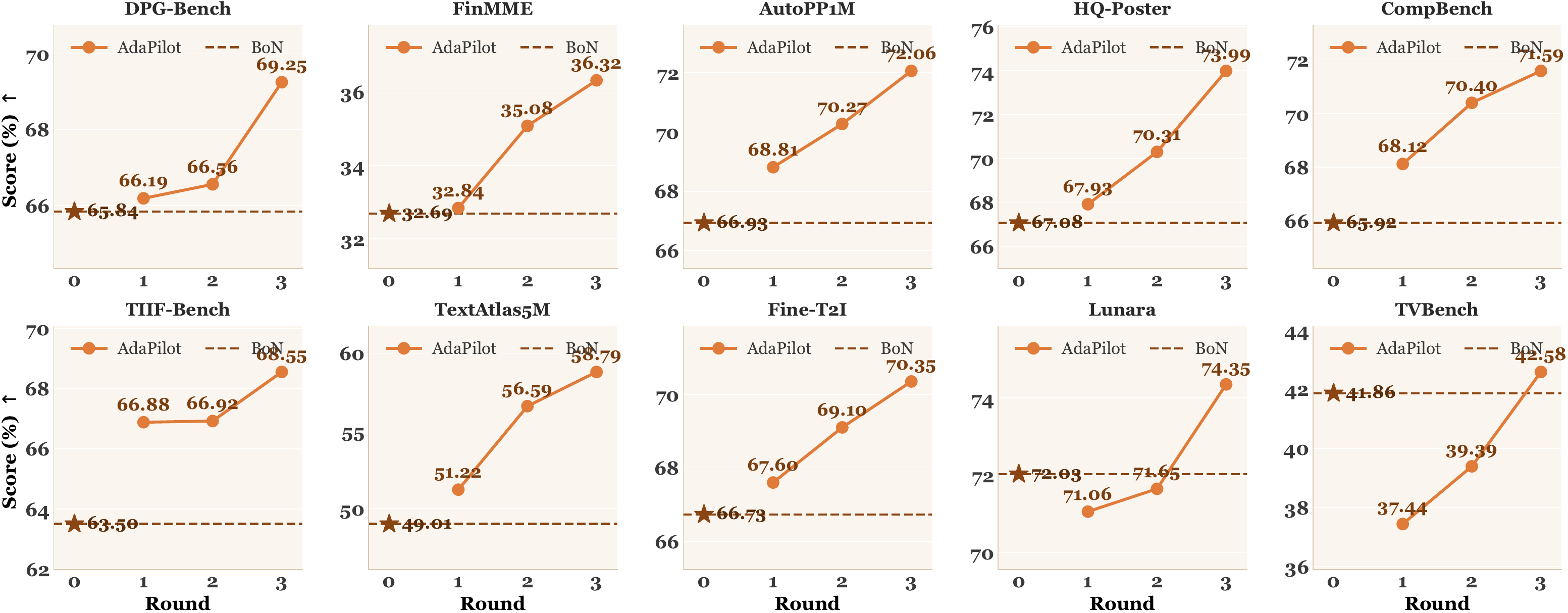}
\caption{Seven-metric averages for three of five rounds. Stars denote pre-feedback Round 0 and dashed lines denote BoN.}
\label{fig:round}
\end{figure*}

\subsection{Out-of-Distribution Generalization (RQ2)}
\label{sec:exp_ood}

As shown in Table~\ref{tab:main-ood}, AdaPilot outperforms all baselines on every reported dimension across the OOD datasets. The results illustrate:
\textit{(i)} \textbf{Generalization extends to unseen tasks.} AdaPilot leads across dense-prompt, table, compositional, and instruction-following scenes, while Flow-GRPO and T2I-R1 fall below the Qwen-Image baseline on most OOD metrics.
\textit{(ii)} \textbf{Held-out dimensions also benefit.} AdaPilot achieves substantial gains in Realism and Physics, both excluded from the training reward, showing benefits beyond directly optimized dimensions.
\textit{(iii)} \textbf{Complex task demands benefit consistently.} AdaPilot improves key quality dimensions on TIIF-Bench and CompBench, supporting adaptation to instruction-following and compositional constraints.

\subsection{Cross-Generator Transfer (RQ3)}
\label{sec:exp_transfer}

As shown in Figures~\ref{fig:ood_radar} and~\ref{fig:id_avg}, AdaPilot transfers directly from its training generator to FLUX.2, LongCat, Ovis, and Z-Image without target-specific retraining. The results demonstrate:
\textit{(i)} \textbf{In-distribution transfer remains positive.} All unseen generators achieve higher in-distribution average scores with AdaPilot than their corresponding base models.
\textit{(ii)} \textbf{Joint generator-and-dataset transfer also holds.} On unseen OOD data, every AdaPilot-driven generator obtains a higher overall average than its baseline.
\textit{(iii)} \textbf{Transfer gains are non-uniform.} Their magnitudes vary across target generators and evaluation settings, indicating broad reuse whose effect depends on the target generator rather than being uniform across deployment conditions, though always positive.

\subsection{Ablation Study (RQ4)}
\label{sec:exp_ablation}

As shown in Table~\ref{tab:ablation}, ablating AdaPilot's key components illustrates that:
\textit{(i)} \textbf{End-to-end policy learning is essential.} The full model leads every metric, while removing RL makes the agent underperform the agent-free setting in OCR, aesthetics, and realism, confirming that policy optimization is necessary for consistent gains.
\textit{(ii)} \textbf{Scene-adaptive reward targets task-specific quality.} Removing it lowers every metric, with pronounced losses in text rendering, aesthetics, and realism, showing task-conditioned feedback aligns optimization with scene-specific quality demands.
\textit{(iii)} \textbf{Monotonic improvement models quality evolution.} Removing it lowers every metric and markedly impairs text rendering, aesthetics, and realism, supporting process-level supervision for sustained quality refinement across turns.

\subsection{Turn-wise Quality Analysis (RQ5)}
\label{sec:exp_turn}

Figure~\ref{fig:round} reports the three rounds shared across datasets under the five-step cap. The results indicate:
\textit{(i)} \textbf{Consistent improvement across turns.} All datasets improve monotonically from Rounds 1--3 and finish above BoN, showing visual feedback sustains quality refinement.
\textit{(ii)} \textbf{Larger gains emerge on complex tasks.} Text-rich and aesthetic-critical datasets show the largest turn-wise gains, highlighting iteration under demanding quality constraints.
\textit{(iii)} \textbf{Later turns recover weaker first-round results.} Although Lunara and TVBench begin below BoN, both improve and exceed it by the final turn, showing later feedback can recover weaker initial results and sustain subsequent quality gains.

\definecolor{avg}{rgb}{0.9961,0.8627,0.8588}

\section{Conclusion}
\label{sec:conclusion}
In this work, we propose AdaPilot, a scene-adaptive policy learning framework for cross-generator image quality optimization. By formulating generation as a visual-feedback MDP, AdaPilot shifts the learning signal from generator parameters to an independent agent policy, unifying scene-adaptive quality assessment, process-level evolution modeling, and shared-observation-space policy-generator decoupling for robust zero-shot transfer to unseen generators. This framework charts a viable path from generator-specific optimization to transferable quality optimization policies.

\newpage
\bibliography{custom}

@String{Computer = "{IEEE} Computer" }

@String{Springer = "Springer-Verlag" }

@article{hao2023optimizing,
  title={Optimizing prompts for text-to-image generation},
  author={Hao, Yaru and Chi, Zewen and Dong, Li and Wei, Furu},
  journal={Advances in Neural Information Processing Systems},
  volume={36},
  pages={66923--66939},
  year={2023}
}

@article{manas2024improving,
  title={Improving text-to-image consistency via automatic prompt optimization},
  author={Ma{\~n}as, Oscar and Astolfi, Pietro and Hall, Melissa and Ross, Candace and Urbanek, Jack and Williams, Adina and Agrawal, Aishwarya and Romero-Soriano, Adriana and Drozdzal, Michal},
  journal={arXiv preprint arXiv:2403.17804},
  year={2024}
}

@inproceedings{yang2024mastering,
  title={Mastering Text-to-Image Diffusion: Recaptioning, Planning, and Generating with Multimodal LLMs.},
  author={Yang, Ling and Yu, Zhaochen and Meng, Chenlin and Xu, Minkai and Ermon, Stefano and Cui, Bin},
  booktitle={ICML},
  pages={56704--56721},
  year={2024}
}

@inproceedings{wu2024self,
  title={Self-correcting llm-controlled diffusion models},
  author={Wu, Tsung-Han and Lian, Long and Gonzalez, Joseph E and Li, Boyi and Darrell, Trevor},
  booktitle={Proceedings of the IEEE/CVF Conference on Computer Vision and Pattern Recognition},
  pages={6327--6336},
  year={2024}
}

@inproceedings{zhuo2025reflection,
  title={From reflection to perfection: Scaling inference-time optimization for text-to-image diffusion models via reflection tuning},
  author={Zhuo, Le and Zhao, Liangbing and Paul, Sayak and Liao, Yue and Zhang, Renrui and Xin, Yi and Gao, Peng and Elhoseiny, Mohamed and Li, Hongsheng},
  booktitle={Proceedings of the IEEE/CVF International Conference on Computer Vision},
  pages={15329--15339},
  year={2025}
}

@inproceedings{wallace2024diffusion,
  title={Diffusion model alignment using direct preference optimization},
  author={Wallace, Bram and Dang, Meihua and Rafailov, Rafael and Zhou, Linqi and Lou, Aaron and Purushwalkam, Senthil and Ermon, Stefano and Xiong, Caiming and Joty, Shafiq and Naik, Nikhil},
  booktitle={Proceedings of the IEEE/CVF Conference on Computer Vision and Pattern Recognition},
  pages={8228--8238},
  year={2024}
}

@article{fan2023dpok,
  title={Dpok: Reinforcement learning for fine-tuning text-to-image diffusion models},
  author={Fan, Ying and Watkins, Olivia and Du, Yuqing and Liu, Hao and Ryu, Moonkyung and Boutilier, Craig and Abbeel, Pieter and Ghavamzadeh, Mohammad and Lee, Kangwook and Lee, Kimin},
  journal={Advances in Neural Information Processing Systems},
  volume={36},
  pages={79858--79885},
  year={2023}
}

@article{um2024minority,
  title={Minority-Focused Text-to-Image Generation via Prompt Optimization},
  author={Um, Soobin and Ye, Jong Chul},
  journal={arXiv preprint arXiv:2410.07838},
  year={2024}
}

@article{ichihara2025evaluation,
  title={Evaluation of best-of-n sampling strategies for language model alignment},
  author={Ichihara, Yuki and Jinnai, Yuu and Morimura, Tetsuro and Ariu, Kaito and Abe, Kenshi and Sakamoto, Mitsuki and Uchibe, Eiji},
  journal={arXiv preprint arXiv:2502.12668},
  year={2025}
}

@article{jiang2024comat,
  title={Comat: Aligning text-to-image diffusion model with image-to-text concept matching},
  author={Jiang, Dongzhi and Song, Guanglu and Wu, Xiaoshi and Zhang, Renrui and Shen, Dazhong and Zong, Zhuofan and Liu, Yu and Li, Hongsheng},
  journal={Advances in Neural Information Processing Systems},
  volume={37},
  pages={76177--76209},
  year={2024}
}

@article{liu2025flow,
  title={Flow-grpo: Training flow matching models via online rl},
  author={Liu, Jie and Liu, Gongye and Liang, Jiajun and Li, Yangguang and Liu, Jiaheng and Wang, Xintao and Wan, Pengfei and Zhang, Di and Ouyang, Wanli},
  journal={arXiv preprint arXiv:2505.05470},
  year={2025}
}

@article{jiang2025t2i,
  title={T2i-r1: Reinforcing image generation with collaborative semantic-level and token-level cot},
  author={Jiang, Dongzhi and Guo, Ziyu and Zhang, Renrui and Zong, Zhuofan and Li, Hao and Zhuo, Le and Yan, Shilin and Heng, Pheng-Ann and Li, Hongsheng},
  journal={arXiv preprint arXiv:2505.00703},
  year={2025}
}

@article{jiang2026genagent,
  title={GenAgent: Scaling Text-to-Image Generation via Agentic Multimodal Reasoning},
  author={Jiang, Kaixun and Wang, Yuzheng and Zhou, Junjie and Li, Pandeng and Liu, Zhihang and Xie, Chen-Wei and Chen, Zhaoyu and Zheng, Yun and Zhang, Wenqiang},
  journal={arXiv preprint arXiv:2601.18543},
  year={2026}
}

@article{guo2025deepseek,
  title={DeepSeek-R1 incentivizes reasoning in LLMs through reinforcement learning},
  author={Guo, Daya and Yang, Dejian and Zhang, Haowei and Song, Junxiao and Wang, Peiyi and Zhu, Qihao and Xu, Runxin and Zhang, Ruoyu and Ma, Shirong and Bi, Xiao and others},
  journal={Nature},
  volume={645},
  number={8081},
  pages={633--638},
  year={2025},
  publisher={Nature Publishing Group UK London}
}

@article{hartwig2025survey,
  title={A survey on quality metrics for text-to-image generation},
  author={Hartwig, Sebastian and Engel, Dominik and Sick, Leon and Kniesel, Hannah and Payer, Tristan and Poonam, Poonam and Glockler, Michael and Bauerle, Alex and Ropinski, Timo},
  journal={IEEE Transactions on Visualization and Computer Graphics},
  year={2025},
  publisher={IEEE}
}

@article{team2025longcat,
  title={Longcat-image technical report},
  author={Team, Meituan LongCat and Ma, Hanghang and Tan, Haoxian and Huang, Jiale and Wu, Junqiang and He, Jun-Yan and Gao, Lishuai and Xiao, Songlin and Wei, Xiaoming and Ma, Xiaoqi and others},
  journal={arXiv preprint arXiv:2512.07584},
  year={2025}
}

@article{cai2025z,
  title={Z-image: An efficient image generation foundation model with single-stream diffusion transformer},
  author={Cai, Huanqia and Cao, Sihan and Du, Ruoyi and Gao, Peng and Hao, Aiming and Hoi, Steven and Hou, Zhaohui and Huang, Shijie and Jiang, Dengyang and Jiang, Yuming and others},
  journal={arXiv preprint arXiv:2511.22699},
  year={2025}
}

@article{wu2025qwen,
  title={Qwen-image technical report},
  author={Wu, Chenfei and Li, Jiahao and Zhou, Jingren and Lin, Junyang and Gao, Kaiyuan and Yan, Kun and Yin, Sheng-ming and Bai, Shuai and Xu, Xiao and Chen, Yilei and others},
  journal={arXiv preprint arXiv:2508.02324},
  year={2025}
}

@misc{blackforestlabs2025flux2,
  title={{FLUX.2}: Frontier Visual Intelligence},
  author={{Black Forest Labs}},
  year={2025},
  howpublished={\url{https://bfl.ai/blog/flux-2}},
  note={Official blog post}
}

@article{cao2025controllable,
  title={Controllable generation with text-to-image diffusion models: A survey},
  author={Cao, Pu and Zhou, Feng and Song, Qing and Yang, Lu},
  journal={IEEE Transactions on Pattern Analysis and Machine Intelligence},
  year={2025},
  publisher={IEEE}
}

@article{wang2025ovis,
  title={Ovis-Image Technical Report},
  author={Wang, Guo-Hua and Cao, Liangfu and Cui, Tianyu and Fu, Minghao and Chen, Xiaohao and Zhan, Pengxin and Zhao, Jianshan and Li, Lan and Fu, Bowen and Liu, Jiaqi and others},
  journal={arXiv preprint arXiv:2511.22982},
  year={2025}
}

@article{chen2026survey,
  title={A survey of multimodal hallucination evaluation and detection},
  author={Chen, Zhiyuan and Min, Yuecong and Zhang, Jie and Yan, Bei and Wang, Jiahao and Wang, Xiaozhen and Shan, Shiguang},
  journal={International Journal of Computer Vision},
  volume={134},
  number={3},
  pages={131},
  year={2026},
  publisher={Springer}
}

@article{wang2025promptenhancer,
  title={Promptenhancer: A simple approach to enhance text-to-image models via chain-of-thought prompt rewriting},
  author={Wang, Linqing and Xing, Ximing and Cheng, Yiji and Zhao, Zhiyuan and Li, Donghao and Hang, Tiankai and Tao, Jiale and Wang, Qixun and Li, Ruihuang and Chen, Comi and others},
  journal={arXiv preprint arXiv:2509.04545},
  year={2025}
}

@article{feng2022training,
  title={Training-free structured diffusion guidance for compositional text-to-image synthesis},
  author={Feng, Weixi and He, Xuehai and Fu, Tsu-Jui and Jampani, Varun and Akula, Arjun and Narayana, Pradyumna and Basu, Sugato and Wang, Xin Eric and Wang, William Yang},
  journal={arXiv preprint arXiv:2212.05032},
  year={2022}
}

@article{chefer2023attend,
  title={Attend-and-excite: Attention-based semantic guidance for text-to-image diffusion models},
  author={Chefer, Hila and Alaluf, Yuval and Vinker, Yael and Wolf, Lior and Cohen-Or, Daniel},
  journal={ACM transactions on Graphics (TOG)},
  volume={42},
  number={4},
  pages={1--10},
  year={2023},
  publisher={ACM New York, NY, USA}
}

@inproceedings{bar2023multidiffusion,
  title={MultiDiffusion: fusing diffusion paths for controlled image generation},
  author={Bar-Tal, Omer and Yariv, Lior and Lipman, Yaron and Dekel, Tali},
  booktitle={Proceedings of the 40th International Conference on Machine Learning},
  pages={1737--1752},
  year={2023}
}

@inproceedings{li2023gligen,
  title={Gligen: Open-set grounded text-to-image generation},
  author={Li, Yuheng and Liu, Haotian and Wu, Qingyang and Mu, Fangzhou and Yang, Jianwei and Gao, Jianfeng and Li, Chunyuan and Lee, Yong Jae},
  booktitle={Proceedings of the IEEE/CVF conference on computer vision and pattern recognition},
  pages={22511--22521},
  year={2023}
}

@article{feng2023layoutgpt,
  title={Layoutgpt: Compositional visual planning and generation with large language models},
  author={Feng, Weixi and Zhu, Wanrong and Fu, Tsu-jui and Jampani, Varun and Akula, Arjun and He, Xuehai and Basu, Sugato and Wang, Xin Eric and Wang, William Yang},
  journal={Advances in Neural Information Processing Systems},
  volume={36},
  pages={18225--18250},
  year={2023}
}

@article{zhang2024realcompo,
  title={Realcompo: Balancing realism and compositionality improves text-to-image diffusion models},
  author={Zhang, Xinchen and Yang, Ling and Cai, Yaqi and Yu, Zhaochen and Wang, Kai-Ni and Xie, Jiake and Tian, Ye and Xu, Minkai and Tang, Yong and Yang, Yujiu and others},
  journal={Advances in Neural Information Processing Systems},
  volume={37},
  pages={96963--96992},
  year={2024}
}

@article{wang2024genartist,
  title={Genartist: Multimodal llm as an agent for unified image generation and editing},
  author={Wang, Zhenyu and Li, Aoxue and Li, Zhenguo and Liu, Xihui},
  journal={Advances in Neural Information Processing Systems},
  volume={37},
  pages={128374--128395},
  year={2024}
}

@inproceedings{yang2024idea2img,
  title={Idea2img: Iterative self-refinement with gpt-4v for automatic image design and generation},
  author={Yang, Zhengyuan and Wang, Jianfeng and Li, Linjie and Lin, Kevin and Lin, Chung-Ching and Liu, Zicheng and Wang, Lijuan},
  booktitle={European Conference on Computer Vision},
  pages={167--184},
  year={2024},
  organization={Springer}
}

@article{black2023training,
  title={Training diffusion models with reinforcement learning},
  author={Black, Kevin and Janner, Michael and Du, Yilun and Kostrikov, Ilya and Levine, Sergey},
  journal={arXiv preprint arXiv:2305.13301},
  year={2023}
}

@article{prabhudesai2023aligning,
  title={Aligning text-to-image diffusion models with reward backpropagation},
  author={Prabhudesai, Mihir and Goyal, Anirudh and Pathak, Deepak and Fragkiadaki, Katerina},
  journal={arXiv preprint arXiv:2310.03739},
  year={2023}
}

@article{clark2023directly,
  title={Directly fine-tuning diffusion models on differentiable rewards},
  author={Clark, Kevin and Vicol, Paul and Swersky, Kevin and Fleet, David J},
  journal={arXiv preprint arXiv:2309.17400},
  year={2023}
}

@inproceedings{wu2024deep,
  title={Deep reward supervisions for tuning text-to-image diffusion models},
  author={Wu, Xiaoshi and Hao, Yiming and Zhang, Manyuan and Sun, Keqiang and Huang, Zhaoyang and Song, Guanglu and Liu, Yu and Li, Hongsheng},
  booktitle={European Conference on Computer Vision},
  pages={108--124},
  year={2024},
  organization={Springer}
}

@article{zhang2024confronting,
  title={Confronting reward overoptimization for diffusion models: A perspective of inductive and primacy biases},
  author={Zhang, Ziyi and Zhang, Sen and Zhan, Yibing and Luo, Yong and Wen, Yonggang and Tao, Dacheng},
  journal={arXiv preprint arXiv:2402.08552},
  year={2024}
}

@inproceedings{deng2024prdp,
  title={Prdp: Proximal reward difference prediction for large-scale reward finetuning of diffusion models},
  author={Deng, Fei and Wang, Qifei and Wei, Wei and Hou, Tingbo and Grundmann, Matthias},
  booktitle={Proceedings of the IEEE/CVF Conference on Computer Vision and Pattern Recognition},
  pages={7423--7433},
  year={2024}
}

@article{xu2023imagereward,
  title={Imagereward: Learning and evaluating human preferences for text-to-image generation},
  author={Xu, Jiazheng and Liu, Xiao and Wu, Yuchen and Tong, Yuxuan and Li, Qinkai and Ding, Ming and Tang, Jie and Dong, Yuxiao},
  journal={Advances in Neural Information Processing Systems},
  volume={36},
  pages={15903--15935},
  year={2023}
}

@inproceedings{liang2024rich,
  title={Rich human feedback for text-to-image generation},
  author={Liang, Youwei and He, Junfeng and Li, Gang and Li, Peizhao and Klimovskiy, Arseniy and Carolan, Nicholas and Sun, Jiao and Pont-Tuset, Jordi and Young, Sarah and Yang, Feng and others},
  booktitle={Proceedings of the IEEE/CVF Conference on Computer Vision and Pattern Recognition},
  pages={19401--19411},
  year={2024}
}

@article{yuan2024self,
  title={Self-play fine-tuning of diffusion models for text-to-image generation},
  author={Yuan, Huizhuo and Chen, Zixiang and Ji, Kaixuan and Gu, Quanquan},
  journal={Advances in Neural Information Processing Systems},
  volume={37},
  pages={73366--73398},
  year={2024}
}

@article{xue2025dancegrpo,
  title={Dancegrpo: Unleashing grpo on visual generation},
  author={Xue, Zeyue and Wu, Jie and Gao, Yu and Kong, Fangyuan and Zhu, Lingting and Chen, Mengzhao and Liu, Zhiheng and Liu, Wei and Guo, Qiushan and Huang, Weilin and others},
  journal={arXiv preprint arXiv:2505.07818},
  year={2025}
}

@inproceedings{liu2026llm,
  title={Llm collaboration with multi-agent reinforcement learning},
  author={Liu, Shuo and Liang, Zeyu and Lyu, Xueguang and Amato, Christopher},
  booktitle={Proceedings of the AAAI Conference on Artificial Intelligence},
  volume={40},
  pages={32150--32158},
  year={2026}
}

@article{ma2026fine,
  title={Fine-T2I: An Open, Large-Scale, and Diverse Dataset for High-Quality T2I Fine-Tuning},
  author={Ma, Xu and Zhang, Yitian and Dong, Qihua and Fu, Yun},
  journal={arXiv preprint arXiv:2602.09439},
  year={2026}
}

@article{hu2024ella,
  title={Ella: Equip diffusion models with llm for enhanced semantic alignment},
  author={Hu, Xiwei and Wang, Rui and Fang, Yixiao and Fu, Bin and Cheng, Pei and Yu, Gang},
  journal={arXiv preprint arXiv:2403.05135},
  year={2024}
}

@article{huang2023t2i,
  title={T2i-compbench: A comprehensive benchmark for open-world compositional text-to-image generation},
  author={Huang, Kaiyi and Sun, Kaiyue and Xie, Enze and Li, Zhenguo and Liu, Xihui},
  journal={Advances in Neural Information Processing Systems},
  volume={36},
  pages={78723--78747},
  year={2023}
}

@article{wei2025tiif,
  title={TIIF-Bench: How Does Your T2I Model Follow Your Instructions?},
  author={Wei, Xinyu and Zhang, Jinrui and Wang, Zeqing and Wei, Hongyang and Guo, Zhen and Zhang, Lei},
  journal={arXiv preprint arXiv:2506.02161},
  year={2025}
}

@article{chen2025postercraft,
  title={Postercraft: Rethinking high-quality aesthetic poster generation in a unified framework},
  author={Chen, SiXiang and Lai, Jianyu and Gao, Jialin and Ye, Tian and Chen, Haoyu and Shi, Hengyu and Shao, Shitong and Lin, Yunlong and Fei, Song and Xing, Zhaohu and others},
  journal={arXiv preprint arXiv:2506.10741},
  year={2025}
}

@article{wang2025textatlas5m,
  title={Textatlas5m: A large-scale dataset for dense text image generation},
  author={Wang, Alex Jinpeng and Mao, Dongxing and Zhang, Jiawei and Han, Weiming and Dong, Zhuobai and Li, Linjie and Lin, Yiqi and Yang, Zhengyuan and Qin, Libo and Zhang, Fuwei and others},
  journal={arXiv preprint arXiv:2502.07870},
  year={2025}
}

@article{liu2025showtable,
  title={ShowTable: Unlocking Creative Table Visualization with Collaborative Reflection and Refinement},
  author={Liu, Zhihang and Bao, Xiaoyi and Li, Pandeng and Zhou, Junjie and Liao, Zhaohe and He, Yefei and Jiang, Kaixun and Xie, Chen-Wei and Zheng, Yun and Xie, Hongtao},
  journal={arXiv preprint arXiv:2512.13303},
  year={2025}
}

@inproceedings{fan2026autopp,
  title={AutoPP: Towards Automated Product Poster Generation and Optimization},
  author={Fan, Jiahao and Qin, Yuxin and Feng, Wei and Chen, Yanyin and Li, Yaoyu and Ma, Ao and Li, Yixiu and Zhuang, Li and Bian, Haoyi and Zhang, Zheng and others},
  booktitle={Proceedings of the AAAI Conference on Artificial Intelligence},
  volume={40},
  pages={3768--3776},
  year={2026}
}

@inproceedings{luo2025finmme,
  title={Finmme: Benchmark dataset for financial multi-modal reasoning evaluation},
  author={Luo, Junyu and Kou, Zhizhuo and Yang, Liming and Luo, Xiao and Huang, Jinsheng and Xiao, Zhiping and Peng, Jingshu and Liu, Chengzhong and Ji, Jiaming and Liu, Xuanzhe and others},
  booktitle={Proceedings of the 63rd Annual Meeting of the Association for Computational Linguistics (Volume 1: Long Papers)},
  pages={29465--29489},
  year={2025}
}

@misc{wang2026moonworkslunaraaestheticdataset,
      title={Moonworks Lunara Aesthetic Dataset}, 
      author={Yan Wang and M M Sayeef Abdullah and Partho Hassan and Sabit Hassan},
      year={2026},
      eprint={2601.07941},
      archivePrefix={arXiv},
      primaryClass={cs.CV},
      url={https://arxiv.org/abs/2601.07941}, 
}

@article{wei2022chain,
  title={Chain-of-thought prompting elicits reasoning in large language models},
  author={Wei, Jason and Wang, Xuezhi and Schuurmans, Dale and Bosma, Maarten and Xia, Fei and Chi, Ed and Le, Quoc V and Zhou, Denny and others},
  journal={Advances in neural information processing systems},
  volume={35},
  pages={24824--24837},
  year={2022}
}

@misc{bai2025qwen25vltechnicalreport,
      title={Qwen2.5-VL Technical Report}, 
      author={Shuai Bai and Keqin Chen and Xuejing Liu and Jialin Wang and Wenbin Ge and Sibo Song and Kai Dang and Peng Wang and Shijie Wang and Jun Tang and Humen Zhong and Yuanzhi Zhu and Mingkun Yang and Zhaohai Li and Jianqiang Wan and Pengfei Wang and Wei Ding and Zheren Fu and Yiheng Xu and Jiabo Ye and Xi Zhang and Tianbao Xie and Zesen Cheng and Hang Zhang and Zhibo Yang and Haiyang Xu and Junyang Lin},
      year={2025},
      eprint={2502.13923},
      archivePrefix={arXiv},
      primaryClass={cs.CV},
      url={https://arxiv.org/abs/2502.13923}, 
}

@inproceedings{chen2025t2i,
  title={T2i-copilot: A training-free multi-agent text-to-image system for enhanced prompt interpretation and interactive generation},
  author={Chen, Chieh-Yun and Shi, Min and Zhang, Gong and Shi, Humphrey},
  booktitle={Proceedings of the IEEE/CVF International Conference on Computer Vision},
  pages={19396--19405},
  year={2025}
}

\clearpage

\appendix

\section*{Appendix}

\section{Theoretical Proof}

\subsection{Proof of Proposition 1}
\label{app:proof1}

\noindent\textbf{Proposition 1.} \textit{Multi-turn visual feedback can achieve higher final image quality than single-turn refinement through iterative defect correction.}
\begin{proof}
Let a trajectory contain $n$ generated images, and let
\begin{equation}
Q_j=\sum_{k\in\mathcal{K}(q)}\bar w_k
r_k(\mathcal{I}_j,q)\in[0,1],
\qquad j\in\{1,\ldots,n\},
\end{equation}
denote the scene-adaptive quality of the $j$-th generated image. Since the final image is returned by the last generation action,
\begin{equation}
Q(\mathcal{I}^*,q)=Q_n.
\end{equation}
All expectations below are taken over $q\sim\mathcal{D}$, generator randomness, and policy randomness.

Let $x_j$ denote the observable policy state after receiving the $j$-th image, comprising the interaction history and the encoded visual observation of $\mathcal{I}_j$. A single-turn policy returns $\mathcal{I}_1$ immediately after its generation, whereas a multi-turn policy may use $x_j$ to select another generation sub-action $a_t^{\mathrm{gen}}$ or the termination action $a_t^{\mathrm{fin}}$.

Let $\Pi_{\mathrm{st}}$ and $\Pi_{\mathrm{mt}}$ denote the classes of single-turn policies and multi-turn feedback policies with interaction budget at most $T_{\max}$, respectively. Any single-turn policy can be reproduced by a multi-turn policy that terminates immediately after generating its first image. Therefore,
\begin{equation}
\Pi_{\mathrm{st}}\subseteq\Pi_{\mathrm{mt}}.
\end{equation}
Let $n_\pi$ be the actual number of images generated by policy $\pi$. The inclusion above gives
\begin{equation}
\sup_{\pi\in\Pi_{\mathrm{mt}}}
\mathbb{E}[Q_{n_\pi}]
\geq
\sup_{\pi\in\Pi_{\mathrm{st}}}
\mathbb{E}[Q_1].
\end{equation}
Thus, visual feedback cannot reduce the optimal expected-quality supremum, since a multi-turn policy can always ignore the feedback and reproduce its single-turn counterpart.

We next consider the case in which visual feedback reveals correctable defects. For each $j<T_{\max}$, let $\mathcal{C}_j$ be a set of correctable-defect states identifiable from $x_j$. Suppose that there exists an executable generation sub-action $c_j(x_j)$ such that, whenever $x_j\in\mathcal{C}_j$,
\begin{equation}
\mathbb{E}\!\left[
Q_{j+1}-Q_j
\mid x_j,\,
a_t^{\mathrm{gen}}=c_j(x_j)
\right]
\geq
\kappa_j(x_j)>0,
\end{equation}
where $\kappa_j(x_j)$ is the conditional expected quality gain in that state. This condition requires improvement only in expectation and does not require every stochastic generation to improve strictly.

Construct a multi-turn extension of any single-turn policy as follows: execute $c_j(x_j)$ when $x_j\in\mathcal{C}_j$; otherwise execute $a_t^{\mathrm{fin}}$ and return the current image. Let $n\leq T_{\max}$ be the final number of images generated by this policy. To handle the process after termination uniformly, define the absorbing quality process
\begin{equation}
\widetilde Q_j=Q_{\min(j,n)},
\qquad j\in\{1,\ldots,T_{\max}\},
\end{equation}
and the corresponding defect potential
\begin{equation}
V_j=1-\widetilde Q_j.
\end{equation}
When $j<n$ and $x_j\in\mathcal{C}_j$, the correction condition yields
\begin{equation}
\mathbb{E}[V_{j+1}-V_j\mid x_j]
\leq-\kappa_j(x_j).
\end{equation}
After termination, $\widetilde Q_{j+1}=\widetilde Q_j$ and hence $V_{j+1}=V_j$. The law of total expectation therefore gives
\begin{equation}
\mathbb{E}[V_{j+1}-V_j]
\leq
-\mathbb{E}\!\left[
\mathbf{1}\{j<n,\,x_j\in\mathcal{C}_j\}
\kappa_j(x_j)
\right].
\end{equation}
Summing over $j=1,\ldots,T_{\max}-1$ and using
\begin{equation}
\widetilde Q_{T_{\max}}=Q_n=Q(\mathcal{I}^*,q)
\end{equation}
yields
\begin{equation}
\begin{aligned}
\mathbb{E}[Q_n-Q_1]
&=\mathbb{E}[V_1-V_{T_{\max}}]\\
&\geq\sum_{j=1}^{T_{\max}-1}
\mathbb{E}\!\left[
\mathbf{1}\{j<n,\,x_j\in\mathcal{C}_j\}
\kappa_j(x_j)
\right]
\geq0.
\end{aligned}
\end{equation}
If at least one correctable-defect state is visited with positive probability, i.e., if there exists some $j<T_{\max}$ such that
\begin{equation}
\Pr\!\left(j<n,\,x_j\in\mathcal{C}_j\right)>0,
\end{equation}
then $\kappa_j(x_j)>0$ makes the right-hand side strictly positive, and consequently
\begin{equation}
\mathbb{E}[Q_n]>\mathbb{E}[Q_1].
\end{equation}
Therefore, for any single-turn policy that produces an identifiable and correctable defect with positive probability, there exists a multi-turn feedback extension using the same initial generation whose expected final image quality is strictly higher. Hence, multi-turn visual feedback can achieve higher final image quality than single-turn refinement through iterative defect correction.
\end{proof}

\subsection{Proof of Proposition 2}
\label{app:proof2}

\noindent\textbf{Proposition 2.} \textit{Scene-adaptive multi-metric supervision can enhance quality optimization across heterogeneous tasks by matching rewards to scene-specific requirements.}
\begin{proof}
Group descriptions with identical quality requirements into the same scene, and let
\begin{equation}
Z=z(q)\in\{1,\ldots,S\},
\qquad p_z=\Pr(Z=z)>0
\end{equation}
denote the scene associated with description $q$ and its probability. Let $\mathcal{K}$ be the set of all quality metrics. The requirement weights for scene $z$ are
\begin{equation}
\alpha_k^{(z)}=
\begin{cases}
\bar w_k^{(z)}, & k\in\mathcal{K}(z),\\
0, & k\notin\mathcal{K}(z),
\end{cases}
\qquad
\sum_{k\in\mathcal{K}}\alpha_k^{(z)}=1.
\end{equation}
For a trajectory $\tau$ terminating at image $\mathcal{I}_n$, define its multi-metric quality vector as
\begin{equation}
\mathbf r(\tau,q)
=\bigl(r_k(\mathcal{I}_n,q)\bigr)_{k\in\mathcal{K}}
\in[0,1]^{|\mathcal{K}|}.
\end{equation}
Its scene-specific quality is
\begin{equation}
Q_n=
\bigl(\boldsymbol{\alpha}^{(Z)}\bigr)^\top
\mathbf r(\tau,q).
\end{equation}
Let $\Pi$ be the class of shared policies acting across all scenes. For any $\pi\in\Pi$, define its expected quality vector in scene $z$ as
\begin{equation}
\mathbf v_\pi^{(z)}
=\mathbb{E}\!\left[
\mathbf r(\tau,q)
\mid Z=z,\,
\tau\sim\pi(\cdot\mid q)
\right],
\end{equation}
and concatenate these vectors across scenes:
\begin{equation}
\mathbf v_\pi=
\bigl(\mathbf v_\pi^{(1)},\ldots,
\mathbf v_\pi^{(S)}\bigr).
\end{equation}
The jointly attainable quality region of shared policies is
\begin{equation}
\mathcal Z=
\overline{\operatorname{conv}}
\left\{\mathbf v_\pi:\pi\in\Pi\right\}
\subseteq[0,1]^{S|\mathcal K|}.
\end{equation}
Randomized mixtures of shared policies realize the corresponding convex combinations. Since all scores are bounded, $\mathcal Z$ is compact, so every linear objective attains its maximum on $\mathcal Z$.

The joint scalarization direction induced by scene-adaptive supervision is
\begin{equation}
\boldsymbol{\omega}_{\mathrm{ada}}
=\bigl(
p_1\boldsymbol{\alpha}^{(1)},\ldots,
p_S\boldsymbol{\alpha}^{(S)}
\bigr).
\end{equation}
Accordingly, the expected quality of policy $\pi$ over the scene distribution is
\begin{equation}
J_{\mathrm{ada}}(\pi)
=\boldsymbol{\omega}_{\mathrm{ada}}^\top
\mathbf v_\pi
=\mathbb{E}_{q\sim\mathcal D,\,
\tau\sim\pi(\cdot\mid q)}[Q_n].
\end{equation}
As a comparison, let $\boldsymbol{\beta}$ be a scene-independent metric weighting satisfying
\begin{equation}
\beta_k\geq0,
\qquad
\sum_{k\in\mathcal K}\beta_k=1.
\end{equation}
Its joint scalarization direction is
\begin{equation}
\boldsymbol{\omega}_{\mathrm{fix}}
=\bigl(
p_1\boldsymbol{\beta},\ldots,
p_S\boldsymbol{\beta}
\bigr).
\end{equation}

Following linear scalarization in multi-objective optimization, define the support function of the joint quality region as
\begin{equation}
h_{\mathcal Z}(\boldsymbol{\omega})
=\sup_{\mathbf v\in\mathcal Z}
\boldsymbol{\omega}^\top\mathbf v.
\end{equation}
Since $\mathcal Z$ is compact, there exist
\begin{equation}
\begin{aligned}
\mathbf v_{\mathrm{ada}}
&\in\arg\max_{\mathbf v\in\mathcal Z}
\boldsymbol{\omega}_{\mathrm{ada}}^\top\mathbf v,\\
\mathbf v_{\mathrm{fix}}
&\in\arg\max_{\mathbf v\in\mathcal Z}
\boldsymbol{\omega}_{\mathrm{fix}}^\top\mathbf v.
\end{aligned}
\end{equation}
These are the supporting points selected by scene-adaptive and fixed supervision, respectively, within the same shared-policy space. By definition of the support function,
\begin{equation}
h_{\mathcal Z}(\boldsymbol{\omega}_{\mathrm{ada}})
=\boldsymbol{\omega}_{\mathrm{ada}}^\top
\mathbf v_{\mathrm{ada}}
\geq
\boldsymbol{\omega}_{\mathrm{ada}}^\top
\mathbf v_{\mathrm{fix}}.
\end{equation}
Define the Bayesian decision regret of fixed supervision relative to the scene-specific quality requirements as
\begin{equation}
\mathcal R_{\boldsymbol{\beta}}
=\boldsymbol{\omega}_{\mathrm{ada}}^\top
\bigl(
\mathbf v_{\mathrm{ada}}-
\mathbf v_{\mathrm{fix}}
\bigr)
\geq0.
\end{equation}
Thus, over the same class of shared policies, the optimal expected-quality supremum under scene-adaptive scalarization is no lower than the value selected by fixed scalarization.

When the two scalarizations select different supporting faces and satisfy
\begin{equation}
\begin{aligned}
\boldsymbol{\omega}_{\mathrm{fix}}^\top
\mathbf v_{\mathrm{fix}}
&\geq
\boldsymbol{\omega}_{\mathrm{fix}}^\top
\mathbf v_{\mathrm{ada}},\\
\boldsymbol{\omega}_{\mathrm{ada}}^\top
\mathbf v_{\mathrm{ada}}
&>
\boldsymbol{\omega}_{\mathrm{ada}}^\top
\mathbf v_{\mathrm{fix}},
\end{aligned}
\end{equation}
fixed supervision reverses the strategy ranking induced by the scene-specific requirements, and $\mathcal R_{\boldsymbol{\beta}}>0$. If either supporting point belongs only to the closure, randomized shared policies can approach it arbitrarily closely, so any strict gap remains positive under a sufficiently accurate approximation. The gap reduces to zero when both scalarizations select the same supporting face.

In summary, scene-adaptive multi-metric supervision cannot reduce the optimal expected-quality supremum over shared policies and yields a strict improvement whenever a fixed metric combination misaligns the strategy ranking with scene-specific requirements. Hence, it can enhance quality optimization across heterogeneous tasks by matching rewards to scene-specific requirements.
\end{proof}

\subsection{Proof of Proposition 3}
\label{app:proof3}

\noindent\textbf{Proposition 3.} \textit{Policy-generator decoupling can enable a single policy to improve outputs from unseen generators zero-shot, without updating either model during transfer.}
\begin{proof}
Let $\mathcal G_{\phi}^{\mathrm{train}}$ be the frozen training generator and $\mathcal G_{\phi}^{\mathrm{infer}}$ a previously unseen generator compatible with the shared interaction protocol: both accept text prompts $p_t$, and their outputs can be processed by the same fixed visual encoder $\mathrm{Enc}_{\mathrm{vis}}$. Fix a common description distribution $q\sim\mathcal D$. During transfer, the same policy $\pi_{\theta^*}$ is reused while both its parameters and those of the inference generator remain fixed. The policy has no direct access to generator identity, architecture, parameters, or gradients; any generator dependence can arise only through visual representations of generated images.

With the shared visual encoder fixed and applied identically to both generators, we characterize transfer through the cross-generator shift between the trajectory distributions induced by the shared interaction protocol, as formalized below.

For any fixed policy $\pi$, extend every early-terminated interaction with an absorbing state to the common finite horizon $T_{\max}$. Let $Y_g^\pi$ be the resulting complete trajectory generated by interaction between $\pi$ and generator $g\in\{\mathrm{train},\mathrm{infer}\}$. It contains the description, states, actions, raw generated images, encoded observations, and final image. Both trajectories are defined on the same measurable space $\mathcal Y$; the subscript $g$ only indexes the inducing generator and is not itself an observed trajectory component. Denote the corresponding trajectory law by
\begin{equation}
\mathbb P_g^\pi=\mathcal L(Y_g^\pi).
\end{equation}

For each trajectory, let $n$ be the number of generated images and define the common terminal-quality functional
\begin{equation}
F(Y_g^\pi)=Q_n\in[0,1],
\end{equation}
using the same evaluation criteria for both generators. The expected final quality of $\pi$ under generator $g$ is
\begin{equation}
J_g(\pi)
=\mathbb E_{\mathbb P_g^\pi}[F(Y)]
=\mathbb E[Q_n],
\end{equation}
where the expectation covers description, policy, and generator randomness. Define total variation by
\begin{equation}
\operatorname{TV}(P,Q)
=\sup_{A\in\mathcal B(\mathcal Y)}
|P(A)-Q(A)|.
\end{equation}
Since $F\in[0,1]$, total-variation duality yields
\begin{equation}
\left|
J_{\mathrm{train}}(\pi)-J_{\mathrm{infer}}(\pi)
\right|
\leq
\operatorname{TV}\!\left(
\mathbb P_{\mathrm{train}}^\pi,
\mathbb P_{\mathrm{infer}}^\pi
\right).
\end{equation}

Let $\pi_{\mathrm{ref}}$ be the reference policy embedded in the same interaction protocol: it submits the original description directly and terminates after receiving the first image. Suppose that the optimized policy has positive expected final-quality gain on the training generator,
\begin{equation}
\Delta_{\mathrm{train}}
=J_{\mathrm{train}}(\pi_{\theta^*})
-J_{\mathrm{train}}(\pi_{\mathrm{ref}})>0.
\end{equation}
Define the trajectory shifts of the optimized and reference policies, respectively, as
\begin{equation}
d_*
=\operatorname{TV}\!\left(
\mathbb P_{\mathrm{train}}^{\pi_{\theta^*}},
\mathbb P_{\mathrm{infer}}^{\pi_{\theta^*}}
\right)
\end{equation}
and
\begin{equation}
d_{\mathrm{ref}}
=\operatorname{TV}\!\left(
\mathbb P_{\mathrm{train}}^{\pi_{\mathrm{ref}}},
\mathbb P_{\mathrm{infer}}^{\pi_{\mathrm{ref}}}
\right).
\end{equation}
The preceding total-variation bound gives
\begin{equation}
J_{\mathrm{infer}}(\pi_{\theta^*})
\geq J_{\mathrm{train}}(\pi_{\theta^*})-d_*
\end{equation}
and
\begin{equation}
J_{\mathrm{infer}}(\pi_{\mathrm{ref}})
\leq J_{\mathrm{train}}(\pi_{\mathrm{ref}})+d_{\mathrm{ref}}.
\end{equation}
Subtracting the second inequality from the first yields
\begin{equation}
J_{\mathrm{infer}}(\pi_{\theta^*})
-J_{\mathrm{infer}}(\pi_{\mathrm{ref}})
\geq
\Delta_{\mathrm{train}}-d_*-d_{\mathrm{ref}}.
\end{equation}
Consequently, whenever
\begin{equation}
\Delta_{\mathrm{train}}>d_*+d_{\mathrm{ref}},
\end{equation}
we have
\begin{equation}
J_{\mathrm{infer}}(\pi_{\theta^*})
>J_{\mathrm{infer}}(\pi_{\mathrm{ref}}),
\end{equation}
so the same fixed policy achieves a strictly positive expected quality gain over direct generation on the unseen generator.

This transfer condition can further be controlled through the turn-wise generator shift over the finite horizon. Let $\mathcal H_t$ be the common set of feasible state--action pairs at turn $t$, and let
\begin{equation}
K_g^t(\cdot\mid s_t,a_t)
\end{equation}
denote the complete environment-response kernel induced by generator $g$ given $(s_t,a_t)\in\mathcal H_t$. The response includes the raw generated image, encoded observation, and corresponding state transition, with an absorbing transition after termination. Suppose that
\begin{equation}
\begin{aligned}
&\sup_{(s_t,a_t)\in\mathcal H_t}
\operatorname{TV}\!\left(
K_{\mathrm{train}}^t(\cdot\mid s_t,a_t),
K_{\mathrm{infer}}^t(\cdot\mid s_t,a_t)
\right)\\
&\hspace{8em}\leq\delta_t,
\qquad \delta_t\in[0,1].
\end{aligned}
\end{equation}
With the same initial distribution and the same fixed policy in both environments, construct a maximal coupling recursively. Conditioned on identical trajectories through turn $t-1$, the policy actions can be coupled identically, and the environment responses at turn $t$ agree with probability at least $1-\delta_t$. The coupling inequality then gives
\begin{equation}
\operatorname{TV}\!\left(
\mathbb P_{\mathrm{train}}^\pi,
\mathbb P_{\mathrm{infer}}^\pi
\right)
\leq
1-\prod_{t=1}^{T_{\max}}(1-\delta_t)
\leq
\sum_{t=1}^{T_{\max}}\delta_t.
\end{equation}
This bound applies to both $\pi_{\theta^*}$ and $\pi_{\mathrm{ref}}$. Hence, a policy-independent sufficient condition is
\begin{equation}
\Delta_{\mathrm{train}}
>
2\left[
1-\prod_{t=1}^{T_{\max}}(1-\delta_t)
\right].
\end{equation}

In summary, when the quality gain on the training generator exceeds the finite-horizon cross-generator trajectory shift, the same policy $\pi_{\theta^*}$ achieves a strictly positive expected quality gain over the reference generation process on an unseen generator while both policy and generator parameters remain fixed. If this condition holds uniformly for a collection of unseen generators, the same policy transfers zero-shot to all of them while preserving its quality gain. Thus, policy-generator decoupling can support zero-shot cross-generator quality optimization without updating either model.
\end{proof}

\begin{figure*}[t]
\centering
\includegraphics[width=\textwidth]{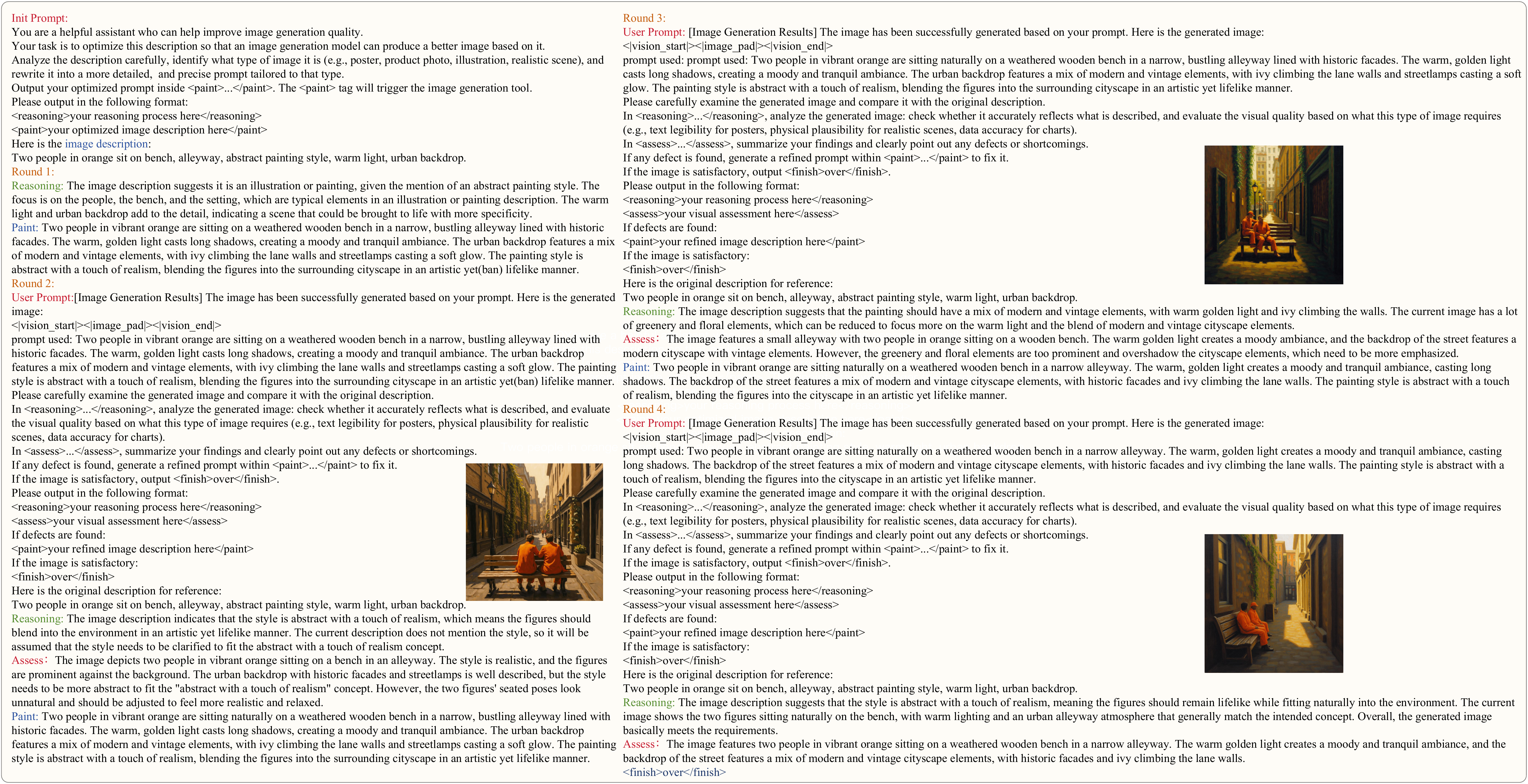}
\caption{An Example of AdaPilot's Multi-turn Visual-Feedback Generation Trajectory.}
\label{fig:full_case}
\end{figure*}

\section{Prompts of AdaPilot}
\label{prompt_details}

Figure~\ref{fig:agent-prompt} presents the task-instruction templates that structure AdaPilot's interaction with the image generator. At the initial turn, the Phase 1 template is combined with the input description to elicit a generation-ready prompt. After each generated image, the Phase 2 template is provided together with the visual result and accumulated interaction context, prompting the agent to assess the current image and either produce a refined prompt or terminate. The interaction also terminates when the turn budget is reached, returning the latest valid image. The structured tags (\texttt{<reasoning>}, \texttt{<paint>}, \texttt{<assess>}, \texttt{<finish>}) instantiate the agent sub-actions and enable format-compliance verification. Figure~\ref{fig:full_case} shows an example trajectory comprising initial prompt optimization, iterative visual assessment and refinement, and final termination.

\section{AdaPilot Algorithm Details}
\label{app:algorithm}

Algorithm~\ref{alg:adapilot} summarizes one policy-update step of AdaPilot. For each description $q$, the behavior snapshot $\pi_{\theta_{\mathrm{old}}}$ samples a group of $N$ complete trajectories against the frozen generator $\mathcal{G}_\phi$. Each sampled trajectory is used directly for reward computation before the group-relative update. The first turn requires a generation action, whereas each later turn requires exactly one generation or termination action, sampled only once. A valid termination action ends the current trajectory, while any output violating this turn-specific constraint is treated as an invalid terminal action and retained for format-compliance evaluation. The three phases in Algorithm~\ref{alg:adapilot} therefore form one complete training update: visual-feedback interaction, AdaReward computation, and end-to-end policy optimization.

\textbf{Complexity Analysis.} Let $L_{i,t}$ and $A_{i,t}$ denote the serialized multimodal context length and the number of agent tokens generated at turn $t$ of trajectory $\tau_i$. Let $C_{\pi}^{\mathrm{dec}}(L_{i,t},A_{i,t})$ denote the autoregressive policy-decoding cost conditioned on that context, and let $C_{\mathcal G}$ and $C_{\mathrm{vis}}$ denote one generator call and one visual encoding, respectively. For a group of $N$ trajectories, the interaction cost is
\begin{equation}
\begin{aligned}
C_{\mathrm{roll}}={}&
\sum_{i=1}^{N}\sum_{t=1}^{T_i}
C_{\pi}^{\mathrm{dec}}(L_{i,t},A_{i,t})\\
&+\sum_{i=1}^{N}\sum_{j=1}^{n_i}
\left(C_{\mathcal G}(p_{i,j})+C_{\mathrm{vis}}(\mathcal{I}_{i,j})\right).
\end{aligned}
\end{equation}
If $C_k(\mathcal{I},q)$ denotes the evaluation cost of metric $k$, the worst-case reward-computation cost is bounded by
\begin{equation}
C_{\mathrm{rew}}=
O\!\left(
\sum_{i=1}^{N}T_i
+\sum_{i=1}^{N}\sum_{j=1}^{n_i}
\sum_{k\in\mathcal{K}(q)}C_k(\mathcal{I}_{i,j},q)
\right).
\end{equation}
For the policy update, the current, behavior, and fixed-anchor policies process the complete serialized multimodal trajectories. With $S_i$ denoting the resulting input sequence for $\tau_i$, the update cost is
\begin{equation}
C_{\mathrm{upd}}=
\sum_{i=1}^{N}\left[
C_{\pi_{\theta_{\mathrm{old}}}}^{\mathrm{fwd}}(S_i)
+C_{\pi_{\theta_0}}^{\mathrm{fwd}}(S_i)
+C_{\pi_\theta}^{\mathrm{fwd+bwd}}(S_i)
\right].
\end{equation}
The response mask restricts loss aggregation to policy-generated positions but does not remove the remaining context from the forward computation. The generator and reward evaluators are frozen, so no gradient is propagated through them, although their inference and evaluation costs remain.

\begin{algorithm*}[!t]
\caption{AdaPilot: Scene-Adaptive Policy Learning for Cross-Generator Image Quality Optimization}
\label{alg:adapilot}
\begin{algorithmic}[1]
\Require Text description $q$, frozen generator $\mathcal{G}_\phi$, behavior policy $\pi_{\theta_{\mathrm{old}}}$, fixed KL anchor $\pi_{\theta_0}$, group size $N>1$, turn budget $T_{\max}\geq1$, reward coefficients $\lambda_Q,\lambda_M$, clip ratio $\epsilon_c$, KL coefficient $\beta_{\mathrm{KL}}$, numerical stabilizer $\epsilon_0>0$
\Ensure Updated policy $\pi_\theta$
\State Set $\theta\leftarrow\theta_{\mathrm{old}}$

\State \textbf{// Phase 1: Multi-Turn Visual-Feedback Interaction}
\For{$i=1$ to $N$}
    \State Initialize $s_{i,1}\leftarrow q$, ordered trajectory $\tau_i\leftarrow\emptyset$, and image count $n_i\leftarrow0$
    \For{$t=1$ to $T_{\max}$}
        \State Sample one structured action $a_{i,t}\sim\pi_{\theta_{\mathrm{old}}}(\cdot\mid s_{i,t})$ and parse its required sub-actions
        \If{$a_{i,t}$ violates the turn-specific output constraint}
            \State Mark $C_{i,t}\leftarrow0$; set $\tau_i\leftarrow\tau_i\mathbin{\oplus}(s_{i,t},a_{i,t})$ and $T_i\leftarrow t$; \textbf{break}
        \EndIf
        \If{$a_{i,t}^{\mathrm{fin}}\in a_{i,t}$}
            \State $\tau_i\leftarrow\tau_i\mathbin{\oplus}(s_{i,t},a_{i,t})$; set $T_i\leftarrow t$; \textbf{break}
        \EndIf
        \State Extract $p_{i,t}$ from $a_{i,t}^{\mathrm{gen}}$ and invoke $\mathcal{G}_\phi$: $n_i\leftarrow n_i+1$; $\mathcal{I}_{i,n_i}\leftarrow\mathcal{G}_\phi(p_{i,t})$
        \State $o_{i,t}\leftarrow(\mathrm{Enc}_{\mathrm{vis}}(\mathcal{I}_{i,n_i}),u_{i,t})$
        \State $\tau_i\leftarrow\tau_i\mathbin{\oplus}(s_{i,t},a_{i,t},o_{i,t})$; set $T_i\leftarrow t$
        \If{$t<T_{\max}$}
            \State $s_{i,t+1}\leftarrow s_{i,t}\oplus a_{i,t}\oplus o_{i,t}$
        \EndIf
    \EndFor
\EndFor

\State \textbf{// Phase 2: Scene-Adaptive Reward Computation (AdaReward)}
\For{$i=1$ to $N$}
    \State Compute $C_{i,t}$ and realized normalized role weights $w_{i,k(t)}$; set $R_{\mathrm{fmt}}(\tau_i)\leftarrow\sum_{t=1}^{T_i}w_{i,k(t)}C_{i,t}$
    \If{$R_{\mathrm{fmt}}(\tau_i)<1$ or $n_i=0$}
        \State $R_i\leftarrow-1+R_{\mathrm{fmt}}(\tau_i)$
    \Else
        \For{$j=1$ to $n_i$}
            \State $Q_{i,j}\leftarrow\sum_{k\in\mathcal{K}(q)}\bar w_k r_k(\mathcal{I}_{i,j},q)$
        \EndFor
        \State $\mathrm{mono}(\tau_i)\leftarrow\frac{Q_{i,n_i}-Q_{i,1}}{\max(1-Q_{i,1},\epsilon_0)}$ if $n_i\geq2$ and $Q_{i,1}<\cdots<Q_{i,n_i}$, otherwise $0$
        \State $R_i\leftarrow\lambda_Q Q_{i,n_i}+\lambda_M\mathrm{mono}(\tau_i)$
    \EndIf
\EndFor

\State \textbf{// Phase 3: End-to-End Policy Optimization (GRPO)}
\State $\mu_q\leftarrow\tfrac{1}{N}\sum_{i=1}^{N}R_i$; $\sigma_q^2\leftarrow\tfrac{1}{N}\sum_{i=1}^{N}(R_i-\mu_q)^2$
\For{$i=1$ to $N$}
    \State $\hat A_i\leftarrow(R_i-\mu_q)/(\sigma_q+\epsilon_0)$ and construct $\mathcal{T}_m^{(i)}$
\EndFor
\State Update $\theta$ by minimizing $\mathcal{L}_{\mathrm{train}}(\theta)$; keep $\pi_{\theta_{\mathrm{old}}}$, $\pi_{\theta_0}$, and $\phi$ fixed
\end{algorithmic}
\end{algorithm*}

At inference, only Phase~1 is executed with the optimized policy $\pi_{\theta^*}$ and the selected frozen generator. Group sampling, reward evaluation, and policy updates are omitted; the latest valid generated image is returned when the policy terminates or the turn budget is reached.

\section{Dataset Details}
\label{app:datasets}
We evaluate AdaPilot on ten public datasets spanning diverse image generation tasks, including compositional generation, dense prompt following, instruction following, fine-grained generation, poster design, text rendering, table visualization, e-commerce product generation, financial chart generation, and aesthetic generation.

\textbf{Fine-T2I}~\citep{ma2026fine} is an open, large-scale, and diverse dataset for text-to-image generation, spanning multiple task combinations, prompt categories, and visual styles. We use its prompts as evaluation inputs to assess generation quality across diverse scenes.

\textbf{DPG-Bench}~\citep{hu2024ella} consists of dense prompts designed to stress-test text-to-image models on dense prompt following, encompassing diverse objects, detailed attributes, complex relationships, and long-text alignment.

\textbf{T2I-CompBench}~\citep{huang2023t2i} is a comprehensive benchmark for open-world compositional text-to-image generation, covering attribute binding, object relationships, and complex compositions.

\textbf{TIIF-Bench}~\citep{wei2025tiif} systematically assesses text-to-image models' ability to follow intricate textual instructions, with prompts organized into multiple difficulty levels, including text rendering and style control evaluation.

\textbf{HQ-Poster-100K}~\citep{chen2025postercraft} is released as part of the PosterCraft framework. This dataset targets high-quality aesthetic poster generation, evaluating text rendering accuracy, layout coherence, and overall visual appeal in poster design scenarios.

\textbf{TextAtlas5M}~\citep{wang2025textatlas5m} provides a large-scale benchmark for dense text image generation, evaluating long-text rendering accuracy and layout complexity across multiple domains.

\textbf{TableVisBench}~\citep{liu2025showtable} contains challenging table visualization instances evaluated across multiple dimensions including data accuracy, text rendering, relative relationship, and aesthetic quality.

\textbf{AutoPP1M}~\citep{fan2026autopp} is a large-scale product poster dataset collected from e-commerce platforms, supporting evaluation of automated product poster generation including background, text, and layout quality.

\textbf{FinMME}~\citep{luo2025finmme} is a financial multi-modal reasoning benchmark encompassing diverse financial domains and chart types. We reconstruct textual descriptions from its financial chart images to create text-to-image generation prompts.

\textbf{Lunara}~\citep{wang2026moonworkslunaraaestheticdataset} is a curated collection of high-quality image-prompt pairs spanning diverse artistic styles, designed for evaluating prompt grounding, style conditioning, and aesthetic alignment in text-to-image generation.

To ensure consistency across datasets and fair training and evaluation, we use Fine-T2I, AutoPP1M, HQ-Poster-100K, Lunara, FinMME, and TextAtlas5M as the six training datasets. We randomly sample 640 instances from each training dataset for training and 128 non-overlapping instances from each of the ten datasets for testing. The training set therefore comprises 3,840 instances, and the test set comprises 1,280 instances.

\begin{figure*}[t]
\centering
\includegraphics[width=\textwidth]{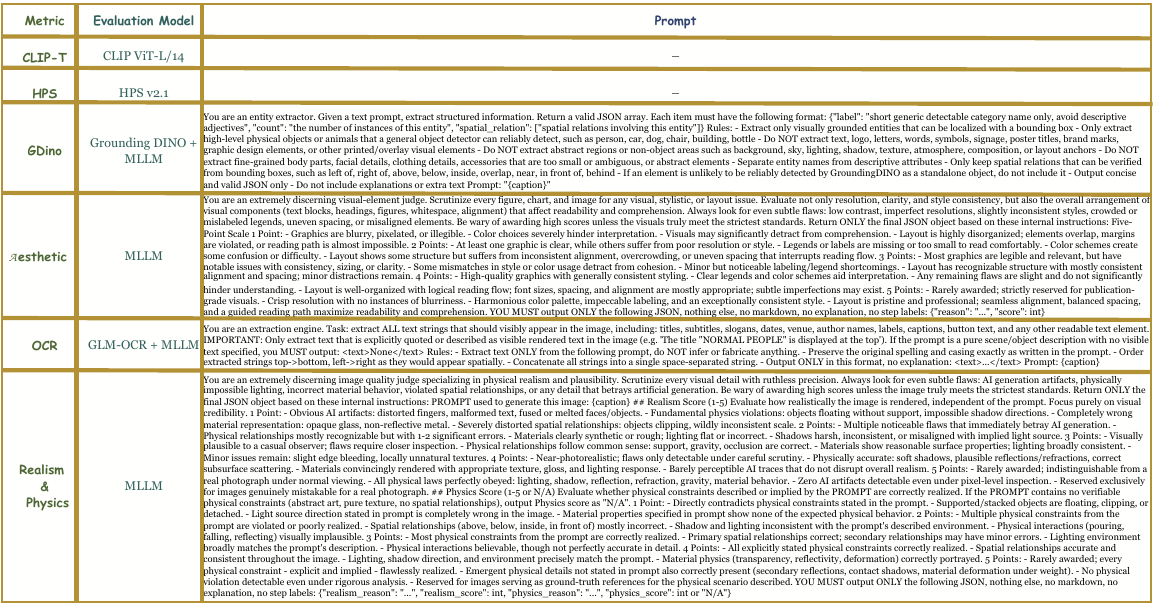}
\caption{Overview of the seven evaluation metrics employed in our proposed AdaPilot, including the evaluation model and prompt for each metric. CLIP-T, HPS, GDino, Aesthetic, and OCR are utilized in the training reward, while Realism and Physics serve as the held-out metrics.}
\label{fig:eval_prompt}
\end{figure*}

\section{Baseline Details}
\label{app:baselines}

We compare AdaPilot with baselines spanning both generator-tuning and generator-frozen paradigms.
For generator-tuning methods, we use the generator and optimization paradigm specified by the corresponding method; frozen-generator baselines retain their reported generator without updating its parameters. The settings are summarized in Table~\ref{tab:hyperparams}.

\textbf{Baseline (Qwen-Image)}~\citep{wu2025qwen} is a text-to-image foundation model from the Qwen family based on the MMDiT architecture, directly generating images from textual descriptions without external optimization.

\textbf{CoT}~\citep{wei2022chain} is a prompting strategy that elicits step-by-step reasoning from large language models before producing the final output, widely adopted in both language and multimodal tasks.

\textbf{SFT} refers to supervised fine-tuning, which updates model parameters on curated input-output pairs through standard cross-entropy loss.

\textbf{GRPO}~\citep{guo2025deepseek} is Group Relative Policy Optimization, a reinforcement learning algorithm that computes relative advantages via group-level reward statistics, eliminating the need for a separate critic model.

\textbf{BoN} (Best-of-N)~\citep{ichihara2025evaluation} is a rejection sampling baseline that generates $N$ images per prompt using the same frozen generator and selects the one with the highest reward score. No training is involved; the method relies entirely on sampling diversity and a reward model for selection, serving as a strong inference-time scaling baseline.

\textbf{Flow-GRPO}~\citep{liu2025flow} applies online policy-gradient reinforcement learning to flow-matching models by converting deterministic ODE sampling into equivalent SDE sampling to introduce stochasticity, combined with a denoising-reduction strategy for efficient GRPO-based generator parameter optimization.

\textbf{T2I-R1}~\citep{jiang2025t2i} introduces collaborative semantic-level and token-level chain-of-thought reasoning into image generation, optimized via BiCoT-GRPO with an ensemble of generation rewards including human preference, object detection, and VQA models.

\textbf{T2I-Copilot}~\citep{chen2025t2i} is a training-free multi-agent system that coordinates three agents---input interpreter, generation engine, and quality evaluator---to automate prompt interpretation, model selection, and iterative refinement for text-to-image generation.

\textbf{MinorityPrompt}~\citep{um2024minority} is a training-free prompt optimization method for text-to-image generation that optimizes prompt embeddings at inference time to steer generation toward low-density regions of the data distribution, improving sample diversity and quality without modifying model parameters.

\begin{table*}[t]
\centering
\begingroup
\fontsize{9}{10.5}\selectfont
\begin{tblr}{
  width=\textwidth,
  colspec={
    Q[l,3.6cm]
    Q[c,1.4cm]
    X[c]
    X[c]
    Q[c,1.7cm]
    Q[c,1.2cm]
    Q[c,1.1cm]
  },
  colsep=0.8mm,
  rowsep=1pt,
  stretch=1.0,
  row{1}={font=\bfseries,halign=c,valign=m},
  hline{1,Z}={1pt},
  hline{2,11}={0.6pt}
}
\textbf{Method} & \textbf{Paradigm} & \textbf{Agent / Model} & \textbf{Generator} & \textbf{Learning Rate} & \textbf{Max Steps} & \textbf{Epochs} \\
Baseline & - & - & Qwen-Image-2512 & - & 1 & - \\
Chain-of-Thought & Frozen & Qwen2.5-VL-7B & Qwen-Image-2512 & - & 1 & - \\
Supervised Fine-Tuning & Frozen & Qwen2.5-VL-7B & Qwen-Image-2512 & $1 \times 10^{-4}$ & 1 & 2 \\
Group Relative Policy Optimization & Frozen & Qwen2.5-VL-7B & Qwen-Image-2512 & $1 \times 10^{-6}$ & 1 & 1 \\
Best-of-N ($N$=5) & Frozen & - & Qwen-Image-2512 & - & 1 & - \\
Flow-GRPO & Tuning & - & Stable Diffusion 3.5 Medium & $1 \times 10^{-4}$ & 1 & - \\
T2I-R1 & Tuning & - & Janus-Pro & $1 \times 10^{-6}$ & 1 & 2 \\
T2I-Copilot & Frozen & Qwen2.5-VL-7B & Qwen-Image-2512 & - & 3 & - \\
MinorityPrompt & Frozen & - & SDXL Lightning & - & $10\,(\mathrm{opt.})$ & - \\
AdaPilot (ours) & Frozen & Qwen2.5-VL-7B & Qwen-Image-2512 & $2 \times 10^{-6}$ & 5 & 1 \\

\end{tblr}
\endgroup
\caption{Hyperparameter settings for AdaPilot and baselines.}
\label{tab:hyperparams}
\end{table*}

\section{Evaluation Metrics}
\label{app:metrics}

We evaluate generated image quality across seven metrics (as illustrated in Figure \ref{fig:eval_prompt}) spanning complementary dimensions. Among these, CLIP-T, HPS, GDino, Aesthetic, and OCR are used in the training reward (AdaReward), while Realism and Physics serve as held-out evaluation dimensions not seen during training.

\textbf{CLIP-T} (Text-Image Semantic Alignment). We compute the cosine similarity between the CLIP ViT-L/14 embeddings of the generated image and the input text description. The raw cosine similarity is linearly normalized to $[0, 1]$:
\begin{equation}
    \text{CLIP-T} = \text{clip}\!\left(\frac{\cos(\mathbf{e}_I,\, \mathbf{e}_T) - 0.15}{0.25},\ 0,\ 1\right),
\end{equation}
where $\mathbf{e}_I$ and $\mathbf{e}_T$ denote the image and text embeddings, respectively. Higher values indicate stronger semantic alignment between the generated image and the textual description.

\textbf{HPS} (Human Preference Score). We adopt HPS v2.1, a scorer trained on large-scale human preference comparison data, which predicts human aesthetic preferences for generated images. We use the official HPS v2.1 scoring procedure and its reported normalization to $[0, 1]$. Higher values indicate greater alignment with human visual preferences.

\textbf{GDino} (Object Detection Accuracy). We use Grounding DINO to evaluate whether objects described in the text prompt are correctly rendered in the generated image. Entities and their attributes are first extracted from the caption via a language model, then detected in the image. The GDino score is a weighted combination of three sub-scores:
\begin{equation}
    \text{GDino} = 0.5 \cdot S_{\text{exist}} + 0.25 \cdot S_{\text{spatial}} + 0.25 \cdot S_{\text{count}},
\end{equation}
where $S_{\text{exist}}$ measures object existence, $S_{\text{spatial}}$ measures spatial relationship accuracy, and $S_{\text{count}}$ measures count accuracy.

\textbf{Aesthetic} (Visual Appeal). A vision-language model evaluates the overall visual quality of the generated image on a five-point scale, assessing resolution, clarity, style consistency, color harmony, layout organization, and labeling quality. The integer score is normalized to $[0, 1]$:
\begin{equation}
    \text{Aesthetic} = \frac{s_{\text{raw}} - 1}{4}, \quad s_{\text{raw}} \in \{1, 2, 3, 4, 5\}.
\end{equation}
\textbf{OCR} (Text Rendering Accuracy). For prompts that require rendered text, a vision-language model extracts the text from the generated image and compares it against the ground-truth text specified in the prompt. Prompts without verifiable rendered text exclude OCR from $\mathcal{K}(q)$ and are not assigned an OCR score. For evaluated prompts, the comparison is performed via character-level alignment using edit distance decomposition into correct matches ($C$), insertions ($I$), deletions ($D$), and substitutions ($S$). The OCR score is computed as the character-level F1-score:
\begin{equation}
    \text{Precision} = \frac{C}{C + I + S}, \quad \text{Recall} = \frac{C}{C + D + S},
\end{equation}
\begin{equation}
    \text{OCR} = \frac{2 \cdot \text{Precision} \cdot \text{Recall}}{\text{Precision} + \text{Recall}}.
\end{equation}
Higher values indicate more accurate text rendering.

\textbf{Realism} (Photographic Realism). A vision-language model evaluates how realistically the image is rendered on a five-point scale, focusing on physical credibility independent of the prompt. The evaluation criteria include AI generation artifacts, material representation, lighting consistency, shadow accuracy, and spatial relationship plausibility. The integer score is normalized to $[0, 1]$:
\begin{equation}
    \text{Realism} = \frac{s_{\text{raw}} - 1}{4}, \quad s_{\text{raw}} \in \{1, 2, 3, 4, 5\}.
\end{equation}
\textbf{Physics} (Physical Plausibility). A vision-language model evaluates whether physical constraints described or implied by the prompt are correctly realized in the generated image, on a five-point scale. The evaluation criteria include spatial relationships, lighting direction, material physics, and physical interactions. For prompts with no verifiable physical constraints, this score is marked as N/A and excluded from the corresponding dataset and overall averages. For evaluated prompts, the integer score is normalized to $[0, 1]$:
\begin{equation}
    \text{Physics} = \frac{s_{\text{raw}} - 1}{4}, \quad s_{\text{raw}} \in \{1, 2, 3, 4, 5\}.
\end{equation}

\begin{figure*}[t]
\centering
\includegraphics[width=\textwidth]{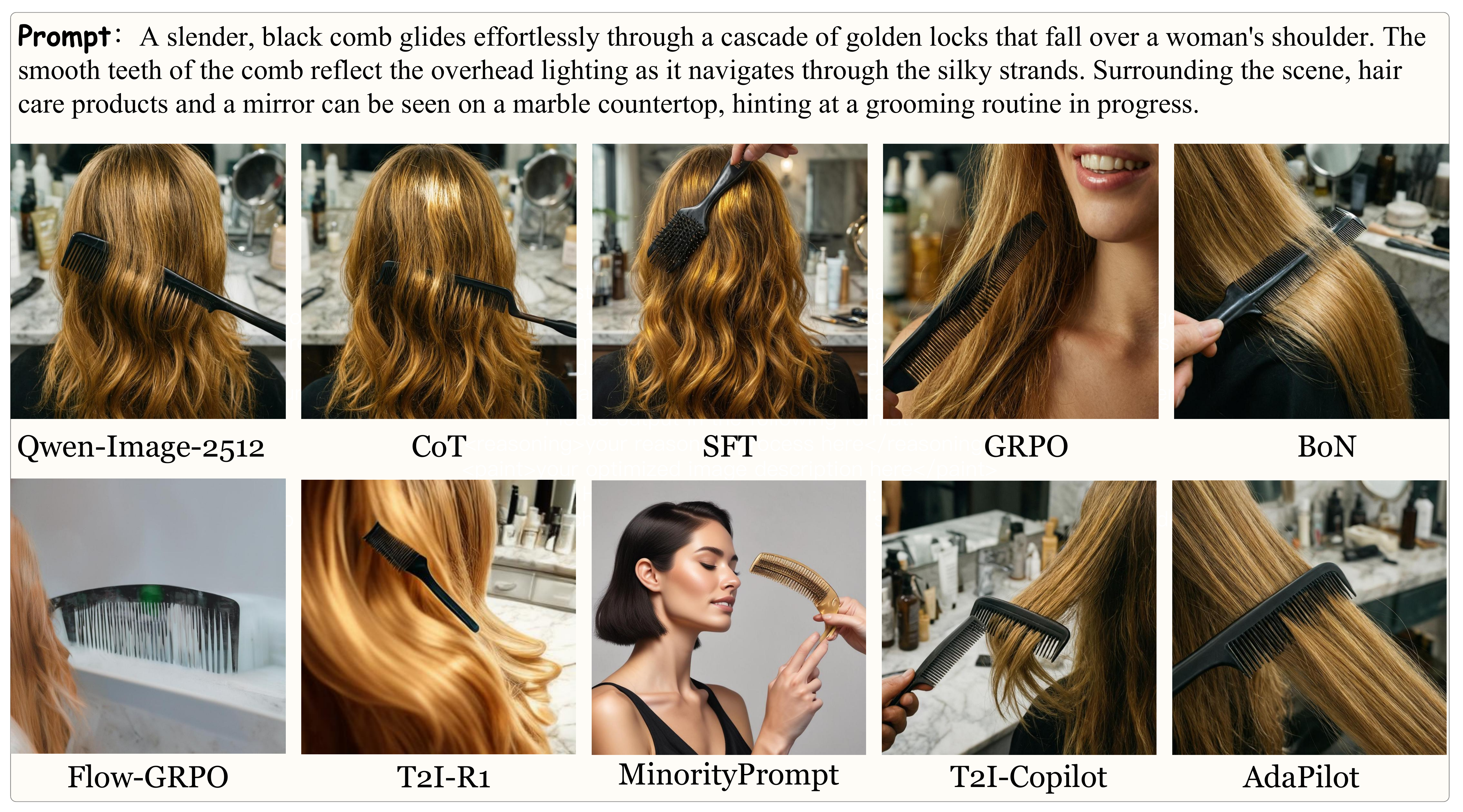}
\caption{Case studies of the baselines and AdaPilot, which include Qwen-Image (baseline generator), CoT, SFT, GRPO, BoN, Flow-GRPO, T2I-R1, T2I-Copilot, MinorityPrompt, and AdaPilot (ours).}
\label{fig:case_study}

\end{figure*}

\section{Implementation Details}
\label{app:implementation}

We adopt Qwen2.5-VL-7B-Instruct as the agent $\mathcal{M}_\theta$, trained with Qwen-Image-2512 serving as the frozen generator $\mathcal{G}_\phi$. All experiments are conducted on two servers, each equipped with eight NVIDIA H100 GPUs (80GB). For reward computation, we deploy CLIP ViT-L/14, HPS v2.1, Grounding DINO, GLM-OCR, and Qwen2.5-VL-72B as evaluators during training. Cross-generator transfer is evaluated zero-shot on FLUX.2, LongCat, Ovis, and Z-Image. For each dataset, all methods are evaluated on the same test set using the same predefined metrics, selected according to the benchmark focus and the availability of verifiable targets. We evaluate AdaPilot and all baselines over three independent runs and report the averaged results. Table~\ref{tab:hyperparams} summarizes the hyperparameter settings for AdaPilot and all baselines.

The relative role weights in $R_{\text{fmt}}$ are set to $v_{\text{first}} = 0.15$, $v_{\text{mid}} = 0.25$, and $v_{\text{last}} = 0.35$, inducing the realized normalized weights $w_{i,k(t)}=v_{k(t)}/\sum_{t'=1}^{T_i}v_{k(t')}$ for trajectory $\tau_i$. Thus, a trajectory receives $R_{\text{fmt}}=1$ if and only if every realized turn is compliant, regardless of its length. Any violation activates the negative-reward branch. The terminal role receives the largest relative weight to emphasize learning autonomous termination. We use $N=4$, $\epsilon_c=0.2$, $\epsilon_0=10^{-6}$, $\beta_{\mathrm{KL}}=0.001$, $\lambda_Q=0.8$, and $\lambda_M=0.2$.

The base weights for CLIP-T, HPS, Aesthetic, GDino, and OCR are $0.15$, $0.20$, $0.20$, $0.25$, and $0.25$, respectively. AdaReward activates all five metrics for HQ-Poster, CLIP-T/HPS/Aesthetic/GDino for Lunara and Fine-T2I, CLIP-T/HPS/GDino for AutoPP1M, and CLIP-T/HPS/Aesthetic/OCR for TextAtlas5M and FinMME. Each active set is renormalized to unit sum, and the mapping remains fixed throughout training.

We optimize AdaPilot with AdamW using a batch size and PPO mini-batch size of $32$. Images are generated at $1024\times1024$ resolution with $50$ sampling steps and a guidance scale of $4.5$. Training uses PyTorch 2.6.0 and vLLM 0.8.4.

\section{Broader Impact}
\label{app:impact}

AdaPilot lowers the cost and expertise barrier for high-quality 
image generation across educational, scientific, and product 
scenes, and its cross-generator transferability extends quality gains 
to open generators evaluated in our experiments, broadening access 
without generator-specific retraining. The same dimensions it improves, 
however, including text rendering, aesthetic fidelity, and physical 
plausibility, are also relevant to document forgery, deceptive 
marketing, brand impersonation, and bias amplification through 
preference-based reward scorers. We mitigate these risks by 
operating strictly through the public prompt interface of existing 
publicly available generators (Qwen-Image, GPT-Image, Gemini, 
etc.) without modifying their parameters or architectures, and by 
accompanying any policy release with usage guidelines prohibiting 
forged-document, impersonation, and identifiable-person generation, 
and recommending provenance signals such as invisible watermarking 
and C2PA-style content authentication for downstream deployments; 
the same scene-adaptive framework can also be reconfigured to 
enforce safety-oriented quality dimensions, which we view as a 
natural follow-up direction. The automated benchmark study uses public 
datasets.

\section{Case Study}
\label{app:case_study}

As shown in Figure~\ref{fig:case_study}, AdaPilot demonstrates more stable scene-adaptive optimization in fine-grained realistic scenarios. This case requires generating a grooming scene where a slender black comb passes through golden hair, while preserving background elements such as the mirror, hair-care products, and marble countertop. Qwen-Image-2512, CoT, and SFT can generate the golden hair and vanity environment, but the interaction between the comb and hair remains weak, lacking the fine-grained action of comb teeth passing through hair strands. GRPO and MinorityPrompt exhibit noticeable semantic shifts in subject composition or object attributes, while Flow-GRPO and T2I-R1 suffer from local structural distortions or unstable object interactions. In contrast, AdaPilot more faithfully preserves the black comb, golden hair, and grooming environment, with the comb teeth better aligned with the hair strands and illumination. The overall result is more natural in terms of object relationships, material texture, and scene coherence, especially under close-up composition. This result suggests that multi-turn visual feedback and scene-adaptive quality optimization help AdaPilot iteratively correct semantic omissions and local defects in intermediate results, thereby improving text faithfulness, object-interaction plausibility, and perceptual realism.

\section{Failure Case Discussion}
\label{app:failure_case}

\begin{figure*}[t]
\centering
\includegraphics[width=\textwidth]{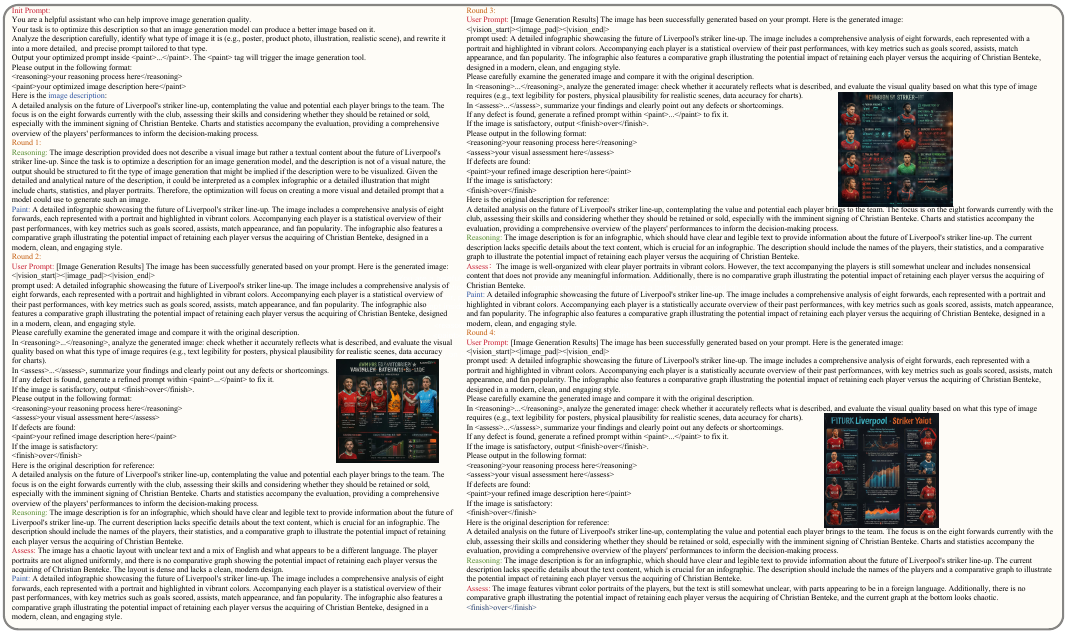}
\caption{Failure case of AdaPilot on a dense sports infographic. Iterative refinement improves layout organization, but the text and charts remain unreliable, and the agent terminates despite identifying residual defects.}
\label{fig:failure_case}
\end{figure*}

As shown in Figure~\ref{fig:failure_case}, this example requests a dense infographic comparing eight Liverpool forwards through portraits, statistics, and a transfer-impact chart. Across rounds, AdaPilot progressively organizes the visual hierarchy and player panels, but the generated text remains largely illegible, player identities and attributes are inconsistent, and the charts do not encode verifiable values. The source description provides neither the players' names nor the underlying statistics, so prompt refinement alone cannot recover the factual content required by the task. Moreover, in the final round, the agent identifies the unresolved text and graph defects but emits the termination token instead of another refinement, revealing a mismatch between visual assessment and the stopping decision. This case exposes two complementary limitations: dense, fact-grounded infographic generation exceeds the generator's reliable text-rendering capacity, while the policy lacks an explicit constraint-verification mechanism for preventing premature termination. Incorporating retrieval-grounded content, structured text and chart rendering, and termination rules tied to verifiable constraints may address these failure modes.

\section{Limitations}
\label{app:limitations}

Despite its strong empirical performance across diverse benchmarks, AdaPilot still has some limitations. First, the multi-turn interaction loop requires sequential image generation at each round, and the quality of later rounds depends critically on the accuracy of earlier quality assessments. If the agent produces an inaccurate quality gap analysis at an early stage, subsequent prompt refinements may drift in a suboptimal direction, and such errors can compound across rounds, degrading final output quality. Second, the scene-adaptive reward mechanism relies on a predefined mapping between scene types and quality dimensions. When encountering scene types outside the current taxonomy, the activated metric subset may not adequately capture the true quality-critical factors, leading to misaligned optimization signals. Third, policy-generator decoupling relies on a shared visual encoder to map outputs from different generators into a common observation space. For generators with distinctive artifacts outside the encoder's pretraining distribution, the encoded observations may not preserve task-relevant quality cues, potentially weakening cross-generator transferability. Finally, automated reward and evaluation models may carry systematic biases and may not fully reflect human judgment.

\section{Future Work}
\label{app:future}

Future work mainly includes the following directions. First, incorporating self-correction mechanisms that allow the agent to detect and recover from erroneous quality assessments during interaction could mitigate the error compounding issue identified above. Second, replacing the predefined scene-metric mapping with an automatic quality dimension discovery module that learns to identify and weight relevant quality dimensions from data would improve generalization to novel scene types without manual metric engineering. Third, extending AdaPilot to other visual generation modalities such as text-to-video and text-to-3D could validate the generality of scene-adaptive policy learning beyond static image synthesis. Fourth, incorporating human preference feedback into the training loop, either through online human evaluation or learned preference models, could further align the optimization signal with perceptual quality and reduce the gap between automated metrics and human judgment.

\clearpage

\end{document}